\documentclass[journal]{IEEEtran}
\usepackage{amsmath,amsfonts}
\usepackage{algorithmic}
\usepackage{algorithm}
\usepackage[caption=false,font=normalsize,labelfont=sf,textfont=sf]{subfig}
\usepackage{subfig}
\usepackage{textcomp}
\usepackage{pdflscape}
\usepackage{stfloats}
\usepackage{url}
\usepackage{verbatim}
\usepackage{graphicx}
\usepackage{bm}
\usepackage{array}
\usepackage{tabularray}
\usepackage{tabularx}
\usepackage{algorithm}
\usepackage{multirow}
\usepackage{booktabs}
\usepackage{array} 
\usepackage{bm}
\usepackage{amssymb}
\usepackage{bbding}
\usepackage{pifont}
\usepackage{wasysym}
\usepackage{amssymb}
\usepackage{graphicx}
\usepackage{makecell}
\usepackage{color}
\usepackage{caption}
\usepackage{xcolor}
\usepackage{multicol}
\usepackage{hyperref}
\hypersetup{hidelinks,
	colorlinks=true,
	allcolors=black,
	pdfstartview=Fit,
	breaklinks=true}
\newcommand{\mysmall}{\fontsize{8.5pt}{2.5pt}\selectfont}
\newcommand{\mysmallsub}{\fontsize{6pt}{2.5pt}\selectfont}

\usepackage[backend=biber, style=ieee]{biblatex}
\ifCLASSINFOpdf
\else
\fi
\begin{document}
%
\title{World Models for Autonomous Driving:\\ An Initial Survey}
%
%
%

\author{Yanchen~Guan$^{*}$,
        Haicheng~Liao$^{*}$,
        Zhenning~Li$^{\dag}$, 
        Jia~Hu$^{\dag}$,
        Runze~Yuan,
        Yunjian~Li,
        Guohui~Zhang, 
       and~Chengzhong~Xu,~\IEEEmembership{Fellow,~IEEE}
\thanks{*\,Authors contributed equally; \dag\,Corresponding author.}
\thanks{Y. Guan, H. Liao, Z. Li, and C. Xu are with the State Key Laboratory of Internet of Things for Smart City, University of Macau, Macau SAR, 999078, China. E-mail: zhenningli@um.edu.mo}
\thanks{J. Hu is with the College of Transportation Engineering, Tongji University, Shanghai, 200092, China. E-mail: hujia@tongji.edu.cn}
\thanks{R. Yuan is with the Department of Automation, Tsinghua University, Beijing, 100084, China.}
\thanks{Y. Li is with the Department of Engineering Science, Macau University of Science and Technology, Macau SAR, 999078, China.}
\thanks{G. Zhang is with the Department of Civil, Environmental and Construction Engineering, University of Hawaii at Manoa, Honolulu, HI 96848, United States.}
\thanks{This research is supported by the Science and Technology Development Fund of Macau SAR (Project no.: 0021/2022/ITP, 001/2024/SKL).}}

\markboth{Journal of IEEE Transactions on Intelligent Vehicless}%
{Guan \MakeLowercase{\textit{et al.}}: Bare Demo of IEEEtran.cls for IEEE Journals}
%



\maketitle

\begin{abstract}
In the rapidly evolving landscape of autonomous driving, the capability to accurately predict future events and assess their implications is paramount for both safety and efficiency, critically aiding the decision-making process. World models have emerged as a transformative approach, enabling autonomous driving systems to synthesize and interpret vast amounts of sensor data, thereby predicting potential future scenarios and compensating for information gaps. This paper provides an initial review of the current state and prospective advancements of world models in autonomous driving, spanning their theoretical underpinnings, practical applications, and ongoing research efforts aimed at overcoming existing limitations. Highlighting the significant role of world models in advancing autonomous driving technologies, this survey aspires to serve as a foundational reference for the research community, facilitating swift access to and comprehension of this burgeoning field, and inspiring continued innovation and exploration. 
\end{abstract}


%
\IEEEpeerreviewmaketitle

\section{Introduction}
The quest to develop autonomous driving systems that seamlessly navigate the intricate tapestry of real-world scenarios remains a formidable frontier in contemporary technology. This challenge is not merely technical but also philosophical, probing the essence of cognition and perception that distinguishes human intelligence from artificial constructs. The crux of this challenge lies in imbuing machines with the kind of intuitive reasoning and `common sense' that humans employ effortlessly. Current machine learning systems, despite their prowess, often stumble in pattern recognition tasks that humans resolve with ease, highlighting a significant gap in our quest for truly autonomous systems \cite{lecun2022path}. On the other hand, human decision-making is deeply rooted in sensory perceptions, constrained by both the memories of these perceptions and direct observations \cite{chang2017code,quiroga2005invariant}. Beyond mere perception, humans possess the uncanny ability to predict the outcomes of their actions, envision potential futures, and anticipate changes in sensory inputs—abilities that underpin our interaction with the world \cite{keller2012sensorimotor}. The endeavor to replicate such capabilities in machines is not just an engineering challenge but a step towards bridging the cognitive divide between human and machine intelligence.

\begin{figure*}[h]
\centering 
\label{Fig.0}
\includegraphics[width=0.7\textwidth]{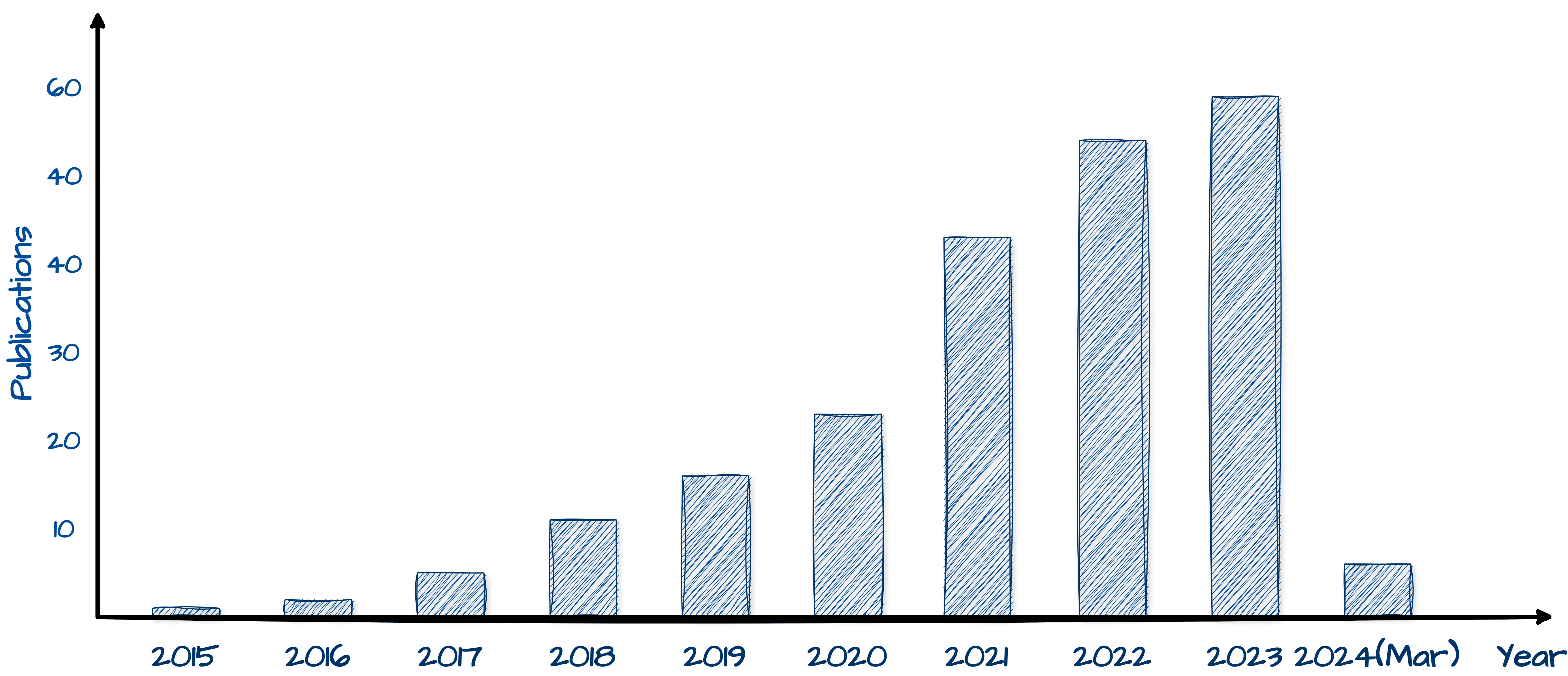}
\caption{Number of Publications related to World Models since 2015. (Data sources: Web of Science Core Collection and Preprint Citation Index.  {Key words: "world model", "world models", "reinforcement learning".})}
\end{figure*}

Addressing this gap, world models have emerged as a critical solution, offering systems the ability to predict and adapt to dynamic environments by emulating human-like perception and decision-making processes. This evolution is essential in the face of real-world scenarios' complexity and unpredictability, where traditional AI approaches struggle to replicate the depth and variability of human cognitive processes. The necessity for world models is underscored by their potential to bridge the cognitive divide between human and machine intelligence, providing a pathway toward more sophisticated autonomous driving systems.

{The journey of world models from conceptual frameworks in control theory during the 1970s to their current prominence in artificial intelligence research reflects a remarkable trajectory of technological evolution and interdisciplinary fusion. The initial formulations in control theory, as laid out by pioneers \cite{rault1978model, bryson2018applied, bryson1996optimal}, were foundational, setting the stage for the integration of computational models in dynamic system management. The underlying logic of Model Predictive Control (MPC) in control theory aligns closely with that of world models \cite{henaff2019model}. Furthermore, the proposal of the mental model systematically described how humans understand the surrounding world and the relationship between abstract concepts, providing theoretical support for world models to imitate humans' understanding of the relationship between the world and themselves \cite{forrester1971counterintuitive,johnson2004history,jones2011mental,johnson1983mental}. These early efforts demonstrated the potential of applying mathematical models to predict and control complex systems, providing guidance for modeling human cognitive systems, which ultimately laid the foundation for the development of world models.

As the field progressed, the advent of neural networks introduced a paradigm shift, allowing for the modeling of dynamic systems with unparalleled depth and complexity \cite{kumpati1990identification,levin1993control,draeger1995model}. This transition from static, linear models to dynamic, non-linear representations facilitated a deeper understanding of environmental interactions, laying the groundwork for the sophisticated world models we see today. The integration of recurrent neural networks (RNNs) was particularly transformative, marking a leap towards systems capable of temporal data processing, essential for predicting future states and enabling abstract reasoning \cite{schmidhuber1990line,schmidhuber2015learning}.

In the early 1990s, using neural networks to learn models for control and applying predictive models in reinforcement learning for both learning and planning ignited pioneers' interest in the construction of learnable models, particularly in the contemplation of models' interactions with the world \cite{narendra1990identification,sutton1991dyna,hunt1992neural,jordan2013forward}. The Dyna architecture elucidated the interaction patterns between agents and the world, as well as the processes of reward and state updates in reinforcement learning \cite{sutton1991dyna}. Research using RNNs to develop internal models for inferring future problems offered a unified framework for the transition from learning to thinking within models \cite{schmidhuber1990making,schmidhuber1990line,chiappa2017recurrent,oh2015action,silver2017predictron,watters2017visual}. Following this, with a long exploration of learning dynamic models and employing these models to train policies, learnable models were back in the center of attention again \cite{deisenroth2011pilco, moerland2023model}. 

 {
In Judea Pearl's hierarchy of causation, the base level is SEEING, which involves identifying associations between elements, a stage where most predictive models currently operate; The middle layer, DOING, refers to actions and interventions, exemplified by the exploration of reinforcement learning; The highest level, IMAGINING, focuses on counterfactual reasoning, understanding the causal relationships between things, enabling the transcendence from complete dependence on datasets \cite{pearl2018book}. This capability allows for dealing with situations beyond the training data through imagination. For learnable models, achieving counterfactual reasoning represents an arduous goal \cite{li2024inferring}. While this ability is innate to humans, crossing this threshold poses a significant challenge for AI. Fortunately, the emergence of world models has introduced a turning point in this dilemma. 
}

The formal unveiling of world models by Ha and Schmidhuber in 2018 was a defining moment that captured the collective aspiration of the AI research community to endow machines with a level of cognitive processing reminiscent of human consciousness \cite{ha2018world,ha2018recurrent}. By harnessing the power of Mixture Density Networks (MDN) and RNNs, this work illuminated the path for unsupervised learning to extract and interpret the spatial and temporal patterns inherent in environmental data. The significance of this breakthrough cannot be overstated, it demonstrated that autonomous systems could achieve a nuanced understanding of their operational environments, predicting future scenarios with an accuracy that was previously unattainable.

In the realm of autonomous driving, the introduction of world models signifies a pivotal shift towards data-driven intelligence, where the capacity to predict and simulate future scenarios becomes a cornerstone for safety and efficiency. The challenge of data scarcity, particularly in specialized tasks such as BEV labeling, underscores the practical necessity of innovative solutions like world models \cite{micheli2022transformers,kaiser2019model,ha2018world}. By generating predictive scenarios from historical data, these models not only circumvent the limitations posed by data collection and labeling but also enhance the training of autonomous systems in simulated environments that can mirror, or even surpass, the complexity of real-world conditions. This approach heralds a new era where autonomous vehicles are equipped with predictive capabilities that reflect a form of intuition, enabling them to navigate and respond to their environment with an unprecedented level of sophistication.

This paper delves into the intricate tapestry of world models, exploring their foundational principles, methodological advancements, and practical applications within the realm of autonomous driving. It navigates through the challenges that beset this field, forecasting future research trajectories and contemplating the broader implications of integrating world models into autonomous systems. In doing so, this work aspires to not only chronicle the progress in this domain but also to inspire a deeper contemplation of the symbiosis between artificial intelligence and human cognition, heralding a new era in autonomous driving technology.

\section{Development of World Models}

This section outlines the intricate architecture of world models, detailing their key components and significant applications across various research studies. These models are engineered to replicate the complex cognitive processes of the human brain, enabling autonomous systems to make decisions and understand their environment in a manner akin to human thinking.

\begin{figure*}[ht]
\centering  
\label{Fig.1}
\includegraphics[width=0.5\textwidth]{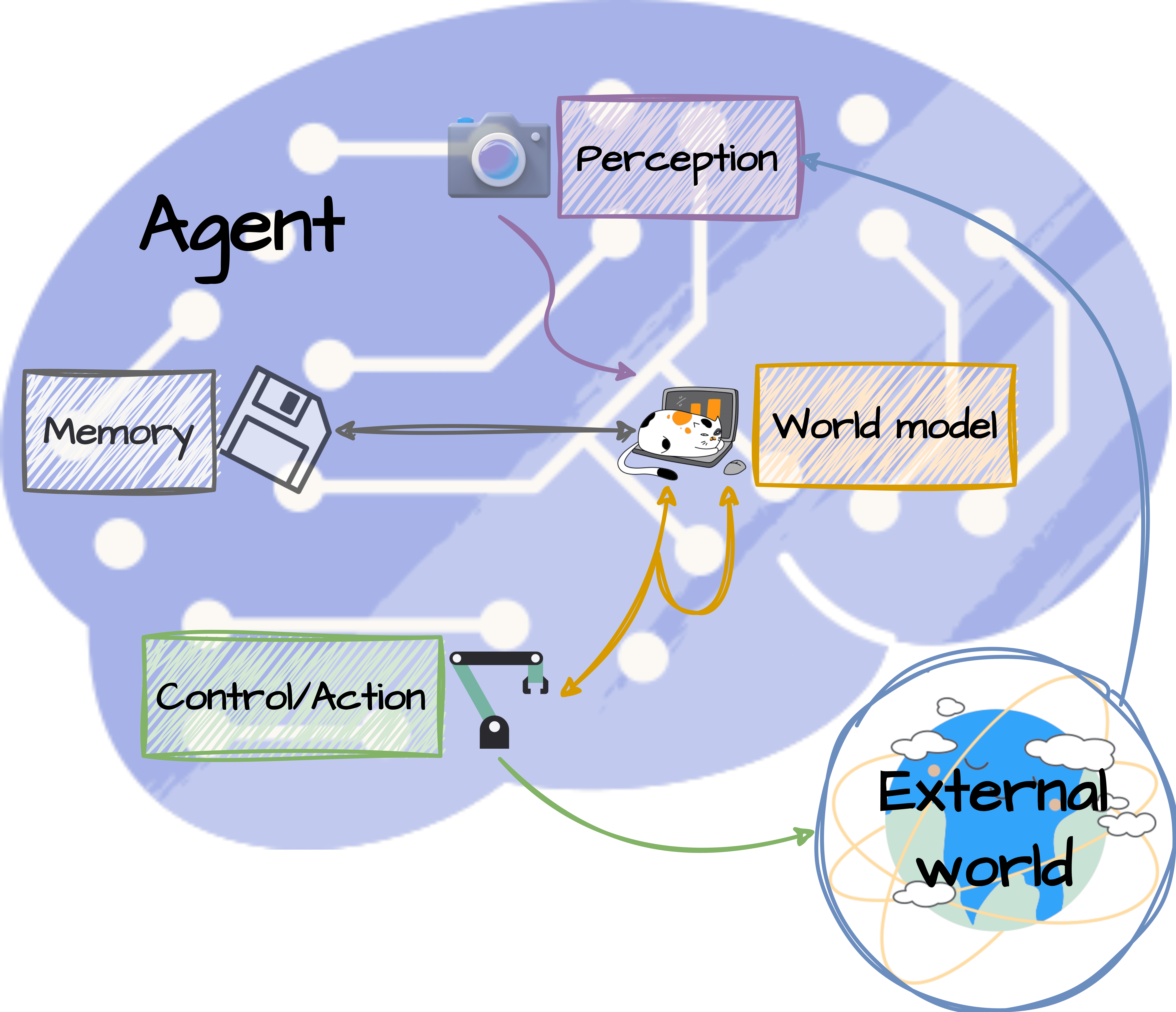}
\caption{Diagram of an Agent's World Model Framework.}
\end{figure*}

\subsection{Architectural Foundations of World Models}\label{World models' structure}
The architecture of a world model is designed to mimic the coherent thinking and decision-making processes of the human brain, integrating several critical components:

\paragraph{Perception Module} This foundational element acts as the system's sensory input, analogous to human senses. It employs advanced sensors and encoder modules, such as Variational Autoencoder (VAE) \cite{kingma2013auto,doersch2016tutorial,rezende2014stochastic}, Masked Autoencoder (MAE) \cite{bardes2023mc,he2022masked}, and Discrete Autoencoder (DAE) \cite{van2017neural,micheli2022transformers}, to process and compress environmental inputs (images, videos, text, control commands) into a more manageable format. The effectiveness of this module is crucial for the accurate perception of complex, dynamic environments, facilitating a detailed understanding that informs the model's subsequent predictions and decisions.

\paragraph{Memory Module} Serving a role similar to the human hippocampus, the memory module is pivotal for recording and managing past, present, and predicted world states along with their associated costs or rewards \cite{lecun2022path}. It supports both short-term and long-term memory functions by replaying recent experiences, a process that enhances learning and adaptation by incorporating past insights into future decisions \cite{foster2017replay}. This module's ability to synthesize and retain crucial information is essential for developing a nuanced understanding of environmental dynamics over time.

\paragraph{Control/Action Module} This component is directly responsible for interacting with the environment through actions. It evaluates the current state and predictions provided by the world model to determine the optimal sequence of actions aimed at achieving specific goals, such as minimizing costs or maximizing rewards. The sophistication of this module lies in its ability to integrate sensory data, memory, and predictive insights to make informed, strategic decisions that navigate the complexities of real-world scenarios.  {This module distinguishes the decision-making process from the intricate world model module and trains it independently using a minimal parameter set. This design allows for the application of more unconventional training methods, such as evolution strategies \cite{rechenberg1978evolutionsstrategien}, to address challenging reinforcement learning tasks where credit assignment poses significant difficulties.}

\paragraph{World Model Module} As the heart of the architecture, the world model module performs two primary functions: estimating any missing information about the current world state and predicting future states of the environment. This dual capability allows the system to generate a comprehensive, predictive model of its surroundings, accounting for uncertainties and dynamic changes. By simulating potential future scenarios, this module enables the system to prepare and adjust its strategies proactively, mirroring the predictive and adaptive thought processes found in human cognition.

Together, these components form a robust framework that empowers world models to simulate cognitive processes and decision-making akin to humans. By integrating these modules, world models achieve a comprehensive and predictive understanding of their environment, which is crucial for the development of autonomous systems capable of navigating and interacting with the complexity of the real world with unprecedented sophistication.

In high-dimensional sensory input scenarios, world models utilize latent dynamical models to abstractly represent observed information, enabling compact forward predictions within a latent state space. These latent states, being significantly more space-efficient than direct predictions of high-dimensional data, facilitate the execution of numerous parallel predictions \cite{hafner2019dream}, thanks to advancements in deep learning and latent variable models. Take, for instance, the ambiguity of a car's direction at an intersection, a scenario emblematic of the inherent unpredictability of real-world dynamics. Latent variables serve as a powerful tool in representing these uncertain outcomes, setting the stage for world models to envision a spectrum of future possibilities grounded in the present state. The crux of this endeavor lies in harmonizing the deterministic aspects of prediction with the intrinsic uncertainty of real-world phenomena, a balancing act central to the efficacy of world models \cite{diehl2023uncertainty, diehl2021umbrella}.

To navigate this challenge, a variety of strategies have been proposed, from introducing uncertainty via a temperature variable \cite{ha2018world} to adopting structured frameworks like the Recurrent State Space Model (RSSM) \cite{hafner2019dream,hafner2019learning,hafner2020mastering,hafner2023mastering,mendonca2021discovering} and the Joint Embedding Predictive Architecture (JEPA) \cite{lecun2022path,fei2023jepa,bardes2023mc,assran2023self}. These methodologies strive to fine-tune the balance between precision and flexibility in prediction. Moreover, leveraging Top-k sampling and transitioning from CNN-based models to transformer architectures \cite{chen2022transdreamer,micheli2022transformers,zhang2023storm}, such as the Transformer State Space Model (TSSM) or Spatial Temporal Patchwise Transformer (STPT), have shown promise in enhancing model performance by better approximating the complexity and uncertainty of the real world. These solutions strive to align the outputs of world models more closely with the probable developments of the real world. This alignment is crucial because the real world, compared to gaming environments, has a much wider range of influencing factors and a greater degree of randomness in future outcomes. Over-reliance on the highest probability predictions may lead to repetitive cycles in long-term forecasting. Conversely, excessive randomness in predictions can lead to absurd futures that significantly diverge from reality. 

 {In particular, the core structures most commonly employed in world model research are RSSM and JEPA:}

\begin{figure*}[ht]
\centering

\subfloat[\mysmall RNN]{

\includegraphics[width=0.25\textwidth]{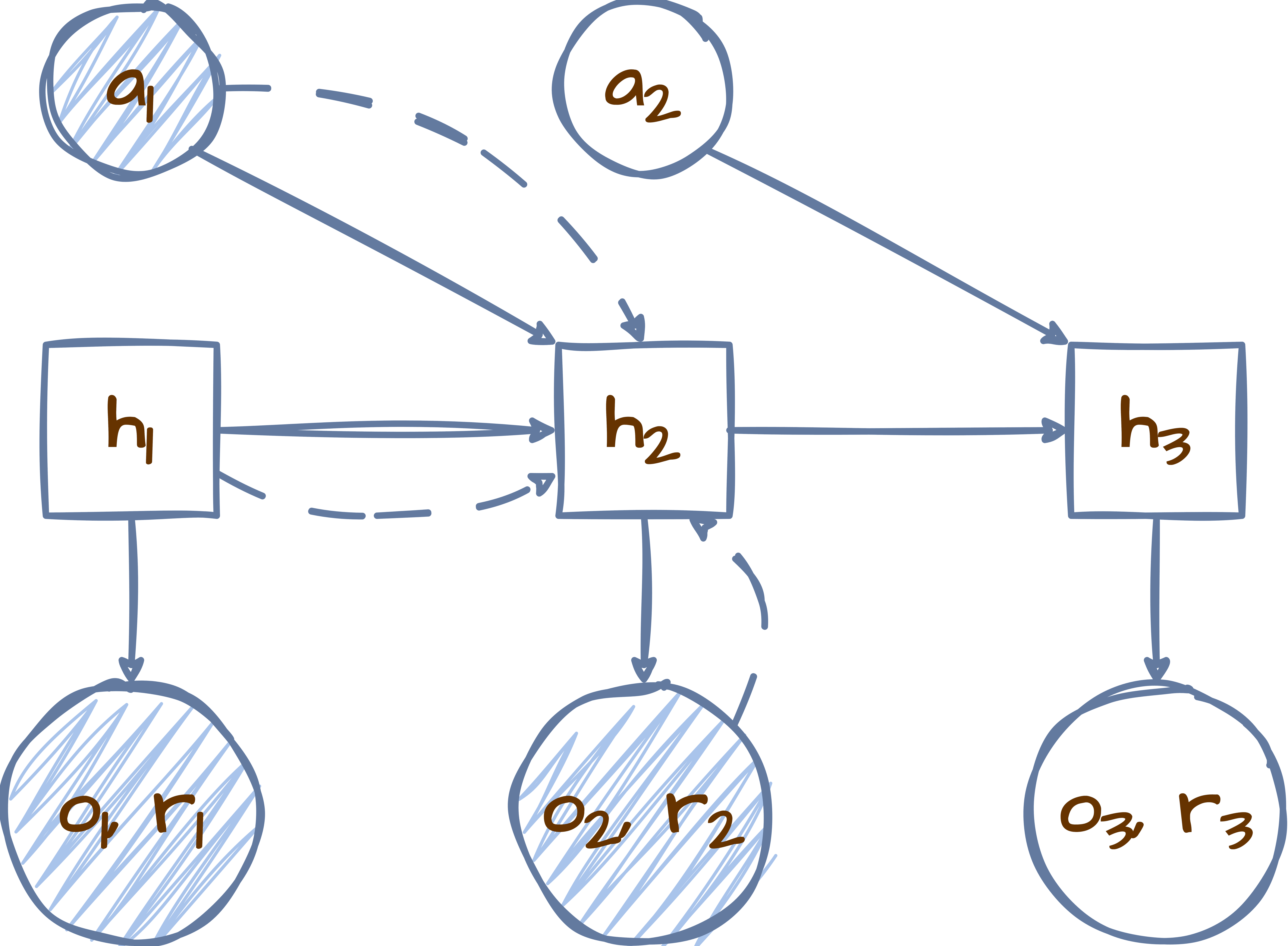}
\label{RNN}}
\hspace{1cm}
\subfloat[\mysmall SSM]{
\label{SSM}
\includegraphics[width=0.25\textwidth]{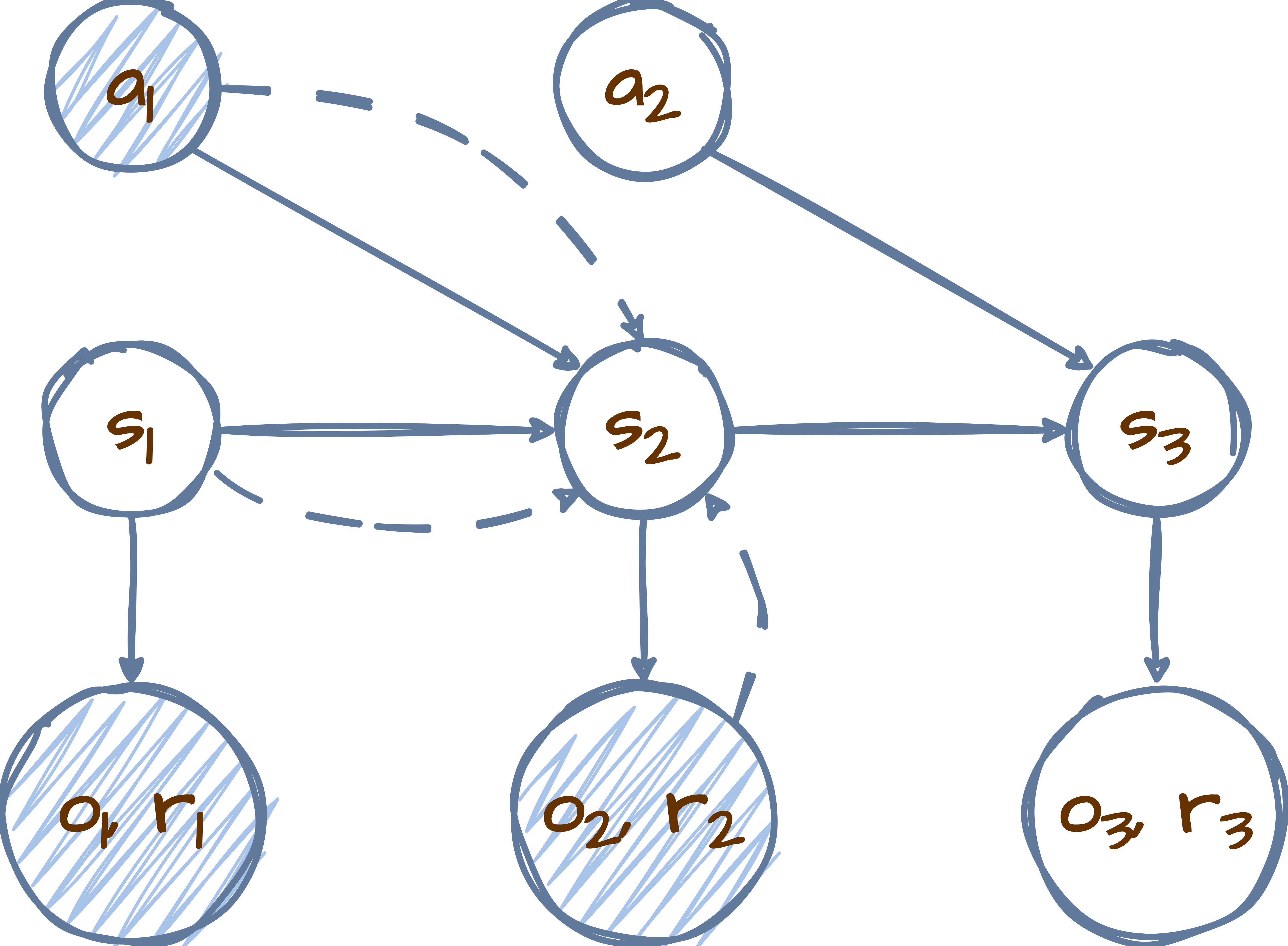}} 
\hspace{1cm}
\subfloat[\mysmall RSSM]{
\label{RSSM}
\includegraphics[width=0.25\textwidth]{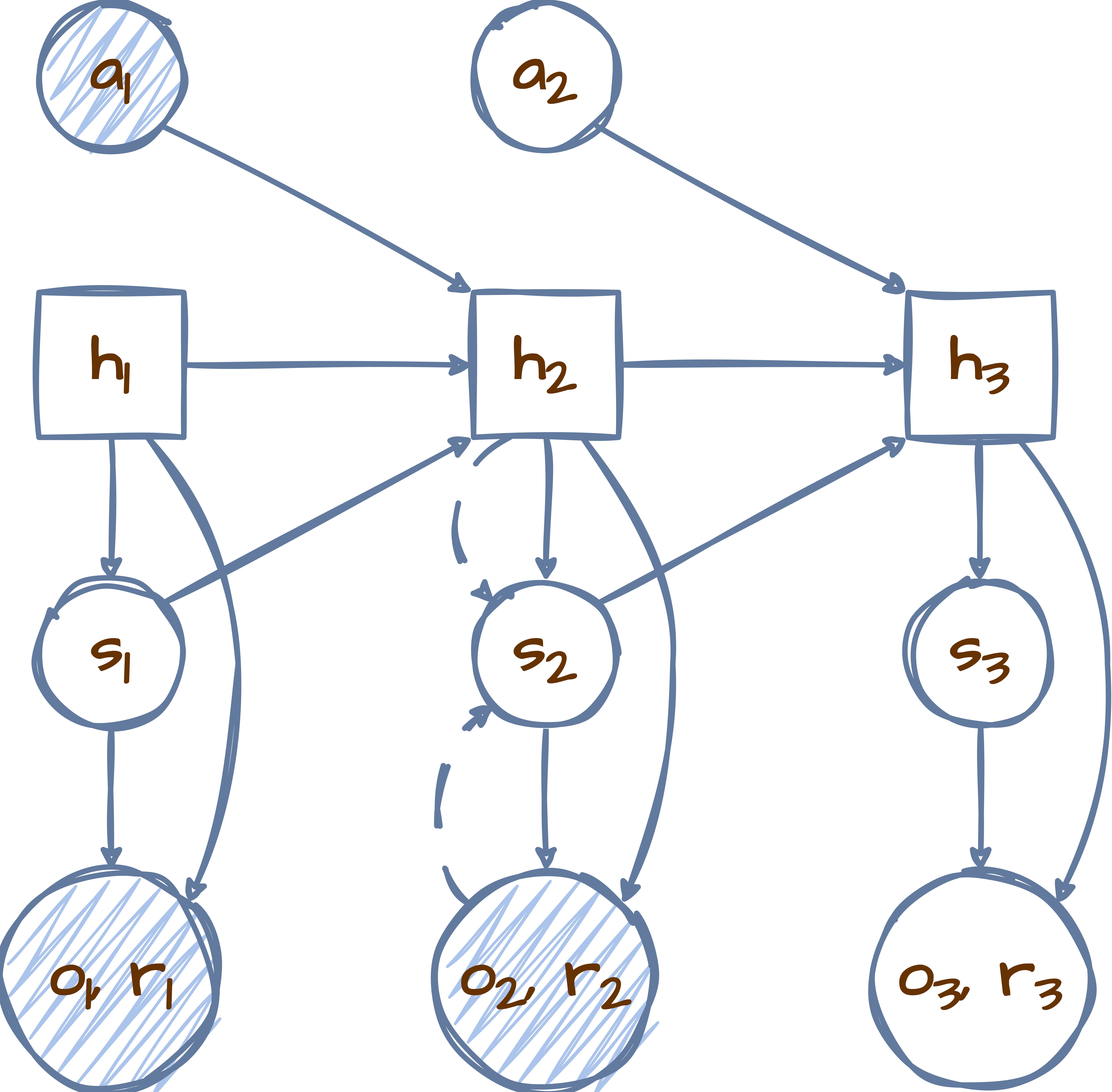}}
  \caption{Comparative Schematic of RNN, SSM, and RSSM Architectures in Latent Dynamics Modeling.}
\label{Fig.2}
\end{figure*}

\noindent\textbf{1) Recurrent State Space Model (RSSM)} \cite{hafner2019learning} stands as a pivotal model within the Dreamer series of world models, designed to facilitate forward prediction purely within the latent space. This innovative structure enables the model to predict through the latent state space, where both stochastic and deterministic paths within the transition model play critical roles in successful planning.

Fig. \ref{Fig.2} illustrates a schematic of the latent dynamics  {models} across three-time steps. Initially observing two-time steps,  {these models then forecast} the third. Here, stochastic variables (circles) and deterministic variables (squares) interact within the models' architecture—solid lines illustrate the generative processes, while dashed lines represent inference pathways. The initial deterministic inference approach in Fig. \ref{RNN} reveals its limitation in capturing diverse potential futures due to its fixed nature. Conversely, an entirely stochastic approach in Fig. \ref{SSM} presents challenges in information retention across time steps, given its inherent unpredictability.

RSSM's innovation lies in its strategic decomposition of states into stochastic and deterministic components in Fig. \ref{RSSM}, effectively harnessing the predictive stability of deterministic elements and the adaptive potential of stochastic elements. This hybrid structure ensures robust learning and prediction capabilities, accommodating the unpredictability of real-world dynamics while preserving information continuity. By marrying the strengths of RNNs with the flexibility of State Space Models (SSM) \cite{hamilton1994state}, RSSM establishes a comprehensive framework for world models, enhancing their ability to predict future states with a balance of precision and adaptability.

We denote the sequence of observations and actions as $\left(\mathbf{x}_0, \boldsymbol{a}_1, \mathbf{x}_1, \boldsymbol{a}_2, \mathbf{x}_2, \ldots, \boldsymbol{a}_T, \mathbf{x}_T\right)$. Namely, the agent takes action $\boldsymbol{a}_{t+1}$ after observing $\mathbf{x}_t$, and receives the next observation $\mathbf{x}_{t+1}$. We omit the reward for simplicity. RSSM models the observations and state transitions through the following generative process:
\begin{equation}
\begin{aligned}
    p\left(\mathbf{x}_{0: T} \mid \boldsymbol{a}_{1: T}\right)=\int \prod_{t=0}^T p\left(\mathbf{x}_t \mid \mathbf{z}_{\leq t}, \boldsymbol{a}_{\leq t}\right) p\left(\mathbf{z}_t \mid \mathbf{z}_{<t}, \boldsymbol{a}_{\leq t}\right) \mathrm{d} \mathbf{z}_{0: T}
\end{aligned}
\end{equation}
where $\mathbf{z}_{0: T}$ are the stochastic latent states. The approximate posterior is defined as:
\begin{equation}
    q\left(\mathbf{z}_{0: T} \mid \mathbf{x}_{0: T}, \boldsymbol{a}_{1: T}\right)=\prod_{t=0}^T q\left(\mathbf{z}_t \mid \mathbf{z}_{<t}, \boldsymbol{a}_{\leq t}, \mathbf{x}_t\right)
\end{equation}

The conditioning on previous states $\mathbf{z}_{<t}$ and actions $\boldsymbol{a}_{\leq t}$ appears multiple times. RSSM uses a shared GRU to compress $\mathbf{z}_{<t}$ and $\boldsymbol{a}_{\leq t}$ into a deterministic encoding $\boldsymbol{h}_t$ :
\begin{equation}
\boldsymbol{h}_t=\operatorname{GRU}\left(\boldsymbol{h}_{t-1}, \operatorname{MLP}\left(\operatorname{concat}\left[\mathbf{z}_{t-1}, \boldsymbol{a}_t\right]\right)\right)
\end{equation}

This is then used to compute the sufficient statistics of the prior, likelihood, and posterior:
\begin{equation}
    \begin{aligned}
p\left(\mathbf{z}_t \mid \mathbf{z}_{<t}, \boldsymbol{a}_{\leq t}\right) & =\operatorname{MLP}\left(\boldsymbol{h}_t\right), \\
p\left(\mathbf{x}_t \mid \mathbf{z}_{\leq t}, \boldsymbol{a}_{\leq t}\right) & =\mathcal{N}\left(\hat{\mathbf{x}}_t, \mathbf{1}\right), \quad \\  \hat{\mathbf{x}}_t & =\operatorname{Decoder}\left(\operatorname{concat}\left[\boldsymbol{h}_t, \mathbf{z}_t\right]\right), \\
q\left(\mathbf{z}_t \mid \mathbf{z}_{<t}, \boldsymbol{a}_{\leq t}, \mathbf{x}_t\right) & =\operatorname{MLP}\left(\operatorname{concat}\left[\boldsymbol{h}_t, \boldsymbol{e}_t\right]\right), \\ \quad \boldsymbol{e}_t & =\operatorname{Encoder}\left(\mathbf{x}_t\right) .
\end{aligned}
\end{equation}

The training objective is to maximize the evidence lower bound (ELBO):
\begin{equation}
\begin{aligned}
    \log p&\left(\mathbf{x}_{0: T} \mid \boldsymbol{a}_{1: T}\right) \geq \mathbb{E}_q\bigg[\sum_{t=0}^T \log p\left(\mathbf{x}_t \mid \mathbf{z}_{\leq t}, \boldsymbol{a}_{\leq t}\right)\\
    &-\mathcal{L}_{\mathrm{KL}}\left(q\left(\mathbf{z}_t \mid \mathbf{z}_{<t}, \boldsymbol{a}_{\leq t}, \mathbf{x}_t\right), p\left(\mathbf{z}_t \mid \mathbf{z}_{<t}, \boldsymbol{a}_{\leq t}\right)\right)\bigg]
\end{aligned}
\end{equation}

\begin{figure*}[ht]

\centering
\subfloat[\mysmallsub Joint-Embedding Architecture]{

\label{Fig.3.sub.1}
\includegraphics[width=0.25\textwidth]{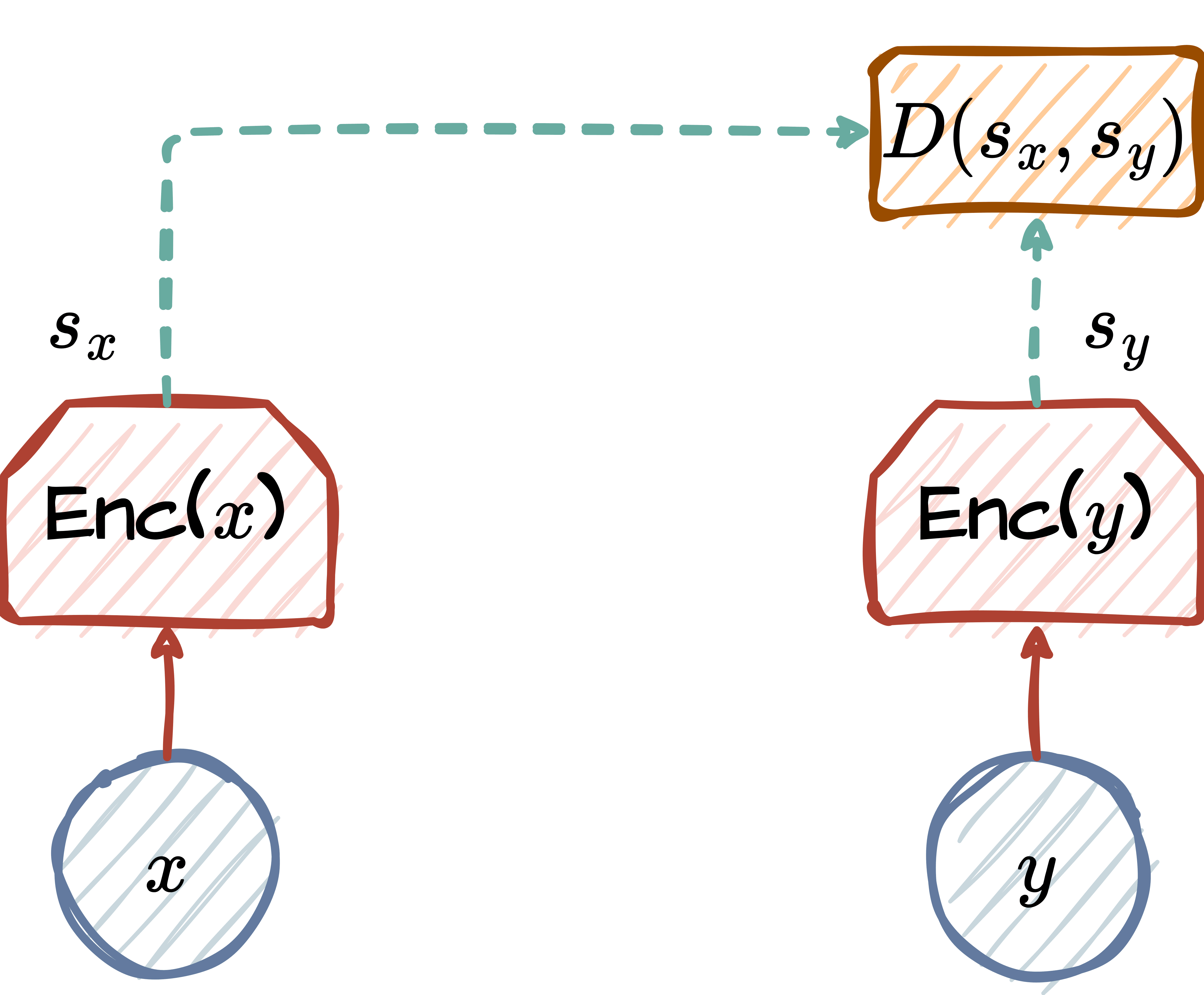}} \hspace{1cm}
\subfloat[\mysmallsub Generative Architecture]{

\label{Fig.3.sub.2}
\includegraphics[width=0.25\textwidth]{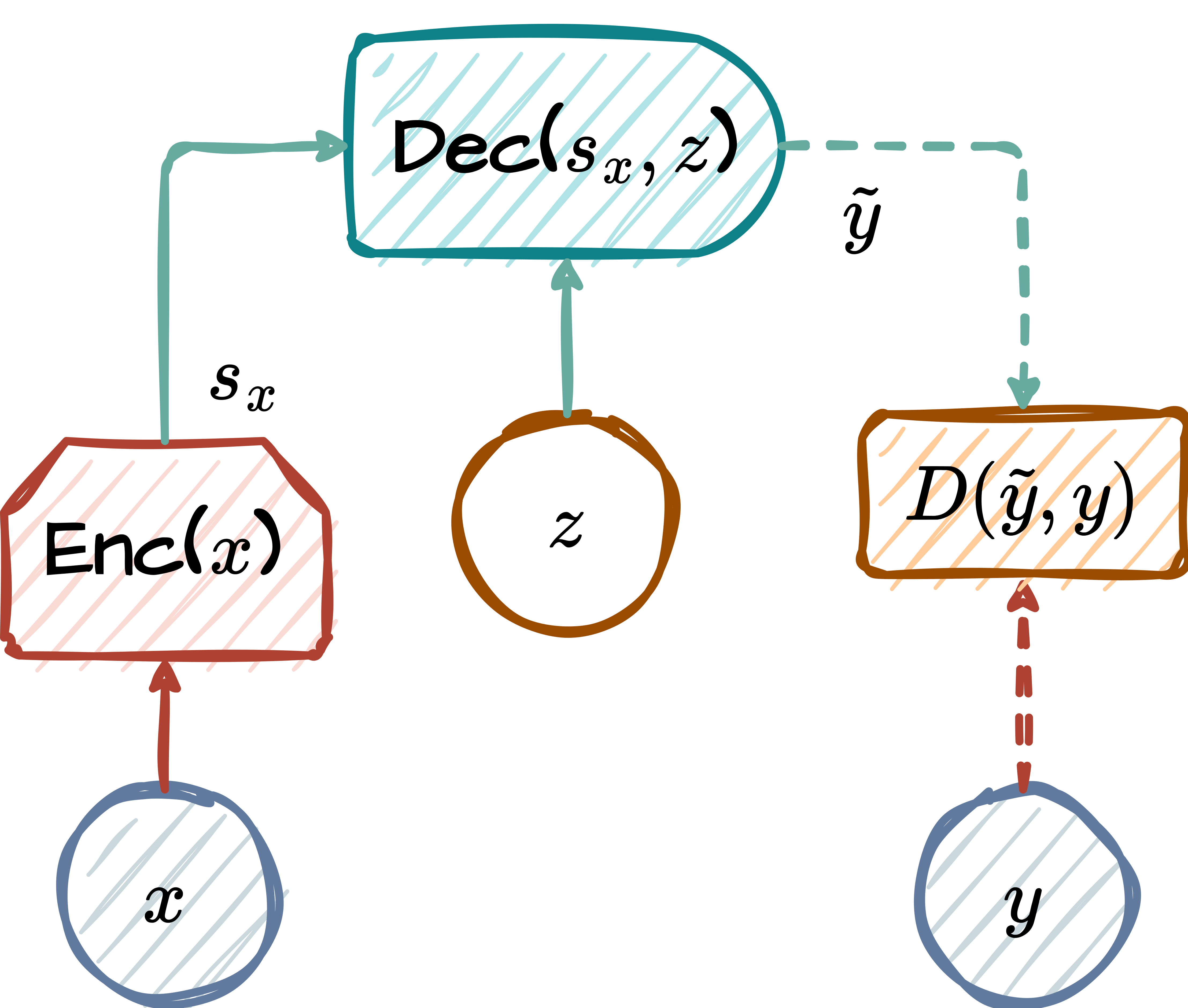}} \hspace{1cm}
\subfloat[\mysmallsub Joint-Embedding Predictive Architecture]{

\label{Fig.3.sub.3}
\includegraphics[width=0.25\textwidth]{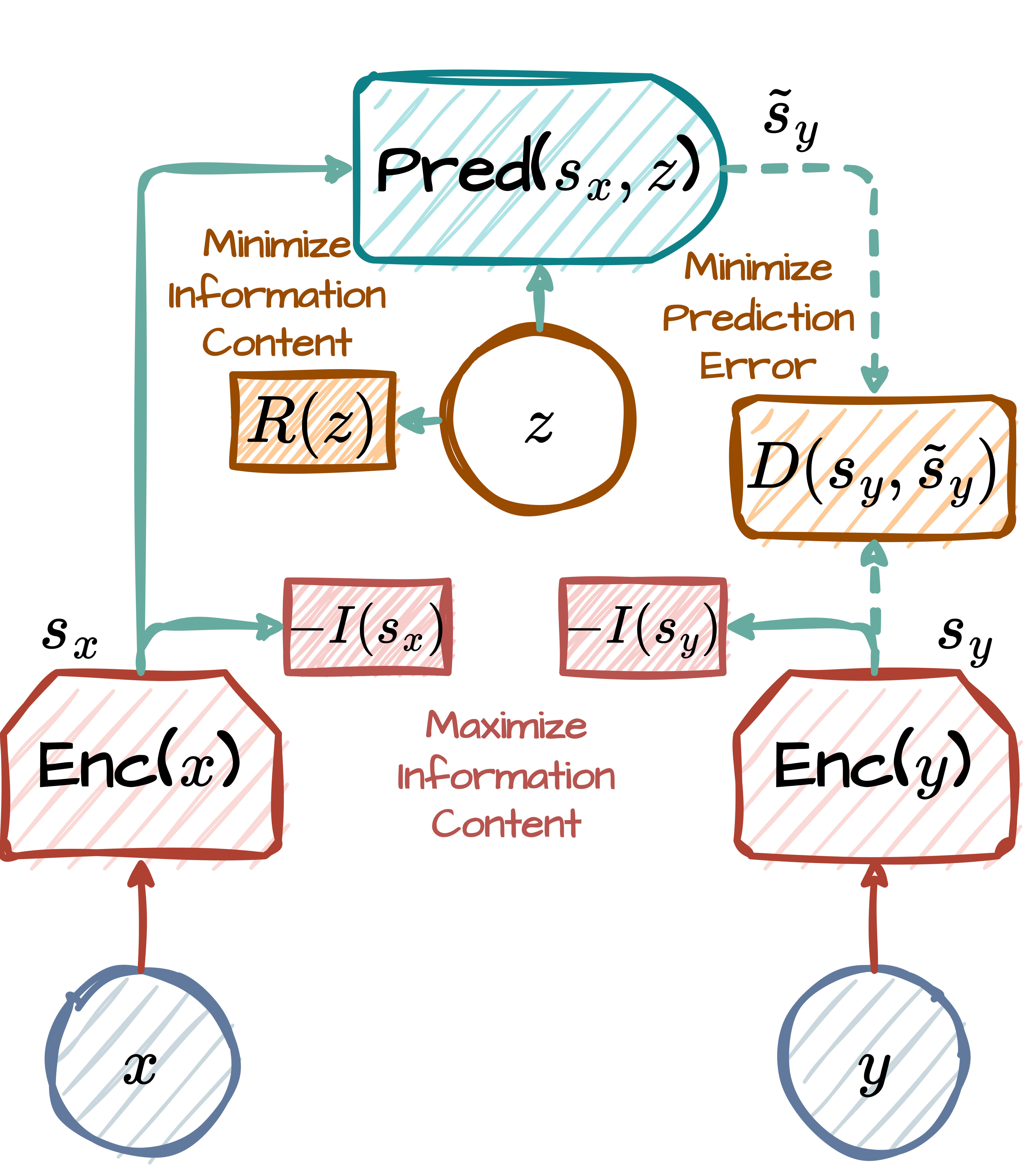}}
  \caption{Comparative Schematic of Joint-Embedding Architecture, Generative Architecture, and Joint-Embedding Predictive Architecture.}
\label{Fig.3}
\end{figure*}

\noindent\textbf{2) Joint-Embedding Predictive Architecture (JEPA) }\cite{lecun2022path} marks a paradigm shift in predictive modeling by focusing on representation space rather than direct, detailed predictions. As shown in  {Fig. \ref{Fig.3}}, by abstracting input (\(\mathbf{x}\)) and target (\(\mathbf{y}\)) through dual encoders into representations (\(\mathbf{s}_x\) and \(\mathbf{s}_y\)), and leveraging a latent variable (\(\mathbf{z}\)) for prediction, JEPA achieves a significant leap in efficiency and accuracy. This model excels in filtering out noise and irrelevancies, concentrating on the essence of the predictive task. The strategic use of the latent variable (\(\mathbf{z}\)) to manage uncertainties further refines the model's focus, enabling it to predict abstract outcomes with heightened precision. By prioritizing relevant features and embracing the inherent uncertainties of predictive tasks, JEPA not only streamlines the prediction process but also ensures outcomes are both relevant and reliable, setting a new standard for predictive modeling in complex environments.

JEPA is underpinned by a multifaceted mathematical model that synthesizes concepts from statistics, machine learning, and optimization. At the heart of JEPA lies the energy function \( \mathbf{E}_w\left(\mathbf{x}, \mathbf{y}, \mathbf{z}; \mathbf{\theta}\right) \), which captures the prediction error within the model. Here, \( \mathbf{x} \) and \( \mathbf{y} \) represent the input and target data, respectively, while \( \mathbf{z} \) stands for the latent variables, and \( \theta \) denotes the model parameters. Mathematically, the energy function \(\mathbf{E}_w\) is defined as:
\begin{equation}
    \mathbf{E}_w\left(\mathbf{x}, \mathbf{y}, \mathbf{z}; \mathbf{\theta}\right)=\Vert\mathbf{s}_y-\operatorname{Pred}\left(\mathbf{s}_x, \mathbf{z}: \phi \right) \Vert_{2}^{2}+\lambda\Vert \mathbf{z} \Vert_{2}^{2}
\end{equation}
Therefore, the core of JEPA's predictive capability is encapsulated in the following key expressions:
1) The squared L2 norm \( \Vert \mathbf{s}_y - \operatorname{Pred}\left(\mathbf{s}_x, \mathbf{z}: \phi \right) \Vert_{2}^{2} \), measures the Euclidean distance between the target representation \( \mathbf{s}_y \) and the predicted representation, highlighting the model's prediction error. The predictive function \(\text{Pred}\) maps the input representation \( \mathbf{s}_x \) and latent variables \( \mathbf{z} \) to the target space, parameterized by \( \phi \); and 2) The regularization term \( \lambda \Vert \mathbf{z} \Vert_2^2 \) penalizes the model's complexity to prevent overfitting by imposing a cost on the magnitude of the latent variables \( \mathbf{z} \), with \( \lambda \) being a non-negative scalar controlling the regularization strength.

Following this, the optimization process aims to minimize \(\mathbf{E}_w\) through finding \(\theta\), \(\phi\), and \(\mathbf{z}\). This can be expressed as a complex Lagrangian optimization problem with constraints on the data distribution:
\begin{equation}
    \operatorname{L}\left( \theta, \phi, \mathbf{z}; \mathbf{x}, \mathbf{y} ,\alpha \right)=\mathbf{E}_w\left(\mathbf{x}, \mathbf{y}, \mathbf{z}; \theta \right)-\alpha\left(h\left(\mathbf{x}, \mathbf{y}, \mathbf{z}; \theta, \phi\right)-\mathbf{c}\right)
\end{equation}
where \(\operatorname{L}\left( \theta, \phi, \mathbf{z}; \mathbf{x}, \mathbf{y},\alpha \right)\) represents the Lagrange equation commonly used in constrained optimization problems, \(\alpha\) is a Lagrange multiplier enforcing the constraint \(h\left(\mathbf{x}, \mathbf{y}, \mathbf{z}; \theta, \phi\right)=\mathbf{c}\), \(h\left(\mathbf{x}, \mathbf{y}, \mathbf{z}; \theta, \phi\right)\) represents the constraint function of the optimization problem parameterized by \(\theta\) and \(\phi\), and \(\mathbf{c}\) is a constant as the target value of function \(h\).

The training of JEPA involves higher-order optimization methods, considering the second-order derivatives to ensure convergence in complex landscapes:
\begin{equation}
    \theta_{\mathbf{t}+1}=\theta_{\mathbf{t}}-\eta\nabla_\theta^2\operatorname{L}\left(\theta_{\mathbf{t}}, \phi_{\mathbf{t}}, \mathbf{z}_{\mathbf{t}}; \mathbf{x}, \mathbf{y}, \alpha_{\mathbf{t}}\right)
\end{equation}
where, \(\theta_{\mathbf{t}+1}\) represents the updated parameter vector at time \(\left(\mathbf{t}+1\right)\). \(\eta\) is the learning rate that determines the step size of the update. \(\eta\nabla_\theta^2\operatorname{L}\left(\theta_{\mathbf{t}}, \phi_{\mathbf{t}}, \mathbf{z}_{\mathbf{t}}; \mathbf{x}, \mathbf{y}, \alpha_{\mathbf{t}}\right)\) represents the Hessian matrix of second-order partial derivatives of the Lagrangian \(\operatorname{L}\) with respect to the parameters \(\theta\) at time \(\mathbf{t}\).

To accommodate the complexity of the optimization landscape, JEPA may utilize higher-order optimization methods that consider second-order derivatives. Furthermore, given the high-dimensional nature of \( \mathbf{z} \) and the possibility of multi-modal distributions, JEPA might employ a variational approximation to the intractable posterior \( p\left(\mathbf{z} \mid \mathbf{x}, \mathbf{y}; \theta\right) \), leading to a variational lower bound:
\begin{equation}\label{log p}
\begin{aligned}
    \log p \left( \mathbf{y}\mid \mathbf{x};\theta,\phi\right)\geq \mathbb{E}_{q\left(\mathbf{z} \mid \mathbf{x}; \psi\right)} \left[\log p\left(\mathbf{y}\mid \mathbf{x},z;\theta,\phi \right)\right]&\\
    -\operatorname{KL}\left[q\left(\mathbf{z} \mid \mathbf{x}; \psi\right)\Vert p\left(\mathbf{z} \mid \mathbf{x}; \theta\right)\right]&
\end{aligned}
\end{equation}
where \( \log p\left(\mathbf{y} \mid \mathbf{x}; \theta, \phi\right) \) is the log-likelihood of the data \( \mathbf{y} \) given \( \mathbf{x} \) under the model parameters \( \theta \) and \( \phi \), \( \mathbb{E}_{q\left(\mathbf{z} \mid \mathbf{x}; \psi\right)}[\cdot] \) is expectation with respect to the distribution \( q\left(\mathbf{z} \mid \mathbf{x}; \psi\right) \), \( \text{KL}\left[q\left(\mathbf{z} \mid \mathbf{x}; \psi\right) \parallel p\left(\mathbf{z} \mid \mathbf{x}; \theta\right)\right] \) is the Kullback-Leibler divergence  between the variational distribution \( q\left(\mathbf{z} \mid \mathbf{x}; \psi\right) \) and the prior distribution \( p\left(\mathbf{z} \mid \mathbf{x}; \theta\right) \).

This inequality, Equation (\ref{log p}), is further used to maximize ELBO to approximate the true posterior distribution.

\begin{table*}[htbp]
\centering
\mysmall
\caption{ {Summary of Recent World Model Applications.}}
\begin{tblr}{
  width = \linewidth,
  colspec = {Q[m, 30]Q[200]Q[242]Q[550]},
  cell{3}{1} = {r=2},
  cell{5}{1} = {r=3},
  cell{8}{1} = {r=2},
  cell{10}{1} = {r=7},
  cell{17}{1} = {r=20},
  cell{37}{1} = {r=5},
  hline{1,42} = {-}{0.08em},
  hline{2-3,5,8,10,17,37} = {-}{0.06em},
}
Year & World Model                              & Core Structure                  & Contribution/Key Idea                                                                                             \\
2018 & World models \cite{ha2018recurrent}      & MDN, RNN                        & Pioneered the first world model for unsupervised learning of spatial and temporal representation of environments.          \\
2019 & PlaNet \cite{hafner2019learning}         & RSSM                            & Introduced a latent dynamics model featuring both deterministic and stochastic transition components for latent prediction.                             \\
     & DreamerV1 \cite{hafner2019dream}         & RSSM                            & Efficient policy learning through the propagation of analytic value gradients for optimized decision-making processes.                                             \\
2020 & DreamerV2 \cite{hafner2020mastering}     & RSSM                            & Utilized discrete multi-class vectors instead of Gaussian distribution for more precise state representation predictions.       \\
     & AWML \cite{kim2020active}                & $\gamma$-Progress               & Developed a world model that employs curiosity-driven exploration.                                                            \\
     & MuZero \cite{schrittwieser2020mastering} & MCTS                            & Integrated a tree-based search with a learned model for improved strategic planning and decision-making.                                                                \\
2021 & EfficientZero \cite{ye2021mastering}     & MCTS                            & Crafted a sample efficient model-based visual RL algorithm built on MuZero.                                               \\
     & Pathdreamer \cite{koh2021pathdreamer}    & RedNet, PatchGAN                & Engineered a world model for Indoor Vision-and-Language Navigation.                                                          \\
2022 & MILE \cite{hu2022model}                  & Probabilistic Generative Model  & Adopted a model-based imitation learning approach for joint policy learning in autonomous driving.                     \\
     & SEM2 \cite{gao2022enhance}               & RSSM                            & Implemented filters to extract key features and balance data distribution, optimizing model performance and accuracy.                                          \\
     & TRANSDREAMER \cite{chen2022transdreamer} & Transformer State-Space Model   & Introduced the first transformer-based stochastic world model.                                                               \\
     & IRIS \cite{micheli2022transformers}      & DAE, Autoregressive Transformer & Constructed a world model combining a discrete autoencoder with an autoregressive transformer.                             \\
     & DreamingV2 \cite{okada2022dreamingv2}    & RSSM                            & Introduced a novel approach by integrating both the discrete representation and the reconstruction-free objective.                                \\
     & DreamerPro \cite{deng2022dreamerpro}     & RSSM, Convolutional Encoder     & Enhanced resistance to visual interference, building upon the Dreamer model's foundation.                                               \\
     & MaxEnt Dreamer \cite{ma2022maxent}       & RSSM, Maximum Entropy           & Implemented maximum entropy reinforcement learning to significantly enhance the world model's exploration capabilities.                 \\
2023 & DriveDreamer \cite{wang2023drivedreamer} & ActionFormer, Diffusion         & Enabled controllable generation of driving scene videos equipped with simplified interactive features.                                          \\
     & Adriver-I \cite{jia2023adriver}          & MLLM, VDM                       & Developed a world model that leverages image-text pairs, combined with large language models.                                                  \\
     & GAIA-1 \cite{hu2023gaia}                 & Autoregressive Transformer      & Crafted a system to generate realistic driving scenes using video, text, and action inputs, providing fine-grained control over the generated environment.                                            \\
     & I-JEPA \cite{assran2023self}             & JEPA                            & Implemented a non-generative approach for self-supervised learning of images.                                                 \\
     & A-JEPA \cite{fei2023jepa}                & JEPA                            & Extended the I-JEPA self-supervised learning approach to the audio spectrum.                                          \\
     & MC-JEPA \cite{bardes2023mc}              & JEPA                            & Employed self-supervised optical flow estimation from videos to jointly learn motion features.                       \\
     & DreamerV3 \cite{hafner2023mastering}     & RSSM                            & Adopted symlog predictions to seamlessly adapt across all ranges using a tailored set of hyperparameters.                                      \\
     & Daydreamer \cite{wu2023daydreamer}       & RSSM                            & Utilized robots to deploy the Dreamer model in real-world settings, enabling direct learning without simulators.                                \\
     & Drive-WM \cite{wang2023driving}          & VAE, Diffusion                  & Introduced the first multi-view, end-to-end autonomous driving world model.                                                   \\
     & TrafficBots \cite{zhang2023trafficbots}  & CVAE, Transformer               & Developed a world model for multimodal motion prediction and end-to-end driving.                                             \\
     & Uniworld \cite{min2023uniworld}          & Transformer                     & Crafted a world model utilizing multi-view, label-free image-LiDAR pairs to model the surrounding scene.                                                                  \\
     & Safe DreamerV3 \cite{huang2023safe}      & RSSM                            & Introduced a world model specifically designed for safe reinforcement learning.                                                                   \\
     & HarmonyDream \cite{ma2023harmony}        & RSSM                            & Incorporated a comprehensive approach by considering the loss of per-dimension scales, dimensions, and training dynamics together.                             \\
     & Dr.G \cite{ha2023dream}                  & RSSM, DCL, RSID                 & Employed a zero-shot MBRL framework that resists visual interference.                                                      \\
     & STORM \cite{zhang2023storm}              & VAE, Transformer                & Combined transformers with variational autoencoders to introduce random noise into model-based reinforcement learning.                                                     \\
    & TWM \cite{robine2023transformer}         & Transformer-XL                  & Utilized the Transformer-XL architecture to learn long-term dependencies while staying computationally efficient. \\
     & DiffDreamer \cite{cai2023diffdreamer}    & Conditional Diffusion Model     & Developed a world model for scene extrapolation.                                                                            \\
     & MUVO \cite{bogdoll2023muvo}              & MLP, GRU                        & Utilized world model to predict 3D occupancy and camera and lidar observations.                                                           \\
     & SWIM \cite{mendonca2023structured}       & RSSM                            & Trained a world model using videos of human hands to enable robots to interact with the environment.                                                                  \\
     & OccWorld \cite{zheng2023occworld}        & S-T Generative Transformer      & Generalized the world model to 3D Occupancy space.                                                                 \\
2024 & WorldDreamer \cite{wang2024worlddreamer} & S-T Patchwise Transformer       & Facilitated learning of dynamic visual signal using attention.                                                     \\
     & S4WM \cite{deng2024facing}               & Parallelizable SSMs             & Modified world model backbones for improving long-term memory.                                                      \\
     & Genie \cite{bruce2024genie}              & ST-Transformer                  & Trained in an unsupervised manner on unlabeled Internet videos to generate interactive environments.
                                        \\
     & V-JEPA \cite{bardes2023v}                & JEPA                            & Extent I-JEPA to videos for self-supervised video pretraining.                                                    \\
     & Think2Drive \cite{li2024think2drive}     & RSSM                            & Efficiently addressed CALAR v2 scenarios with world models.         
\end{tblr}
\label{tab_applications}
\end{table*}

\subsection{Broad Spectrum Applications}\label{Application of World Models}

As shown in TABLE \ref{tab_applications},  {with the evolution of world models, they have been applied across multiple domains, showcasing unparalleled performance in diverse environments,} particularly in gaming, where their capabilities are prominently displayed. In the competitive landscape of the Atari 100k leaderboard, world models dominate, with four out of the top five positions held by these innovative architectures \cite{kaiser2019model, zhang2023storm, micheli2022transformers, ye2021mastering, ma2023harmony}.  {Among these, EfficientZero stands out by significantly enhancing sampling efficiency in image-based reinforcement learning. By leveraging the foundational principles of MuZero and incorporating lightweight improvements, EfficientZero achieved human-comparable gaming proficiency within a mere two hours of training, securing the top position on the leaderboard \cite{schrittwieser2020mastering}.} In the Minecraft game, DreamerV3 marks a milestone as the inaugural model to autonomously mine diamonds, a feat accomplished without leveraging human-generated data or predefined learning curricula. This achievement is attributed to its novel use of Symlog Predictions, facilitating the model's adaptability across diverse environmental scales by employing static symlog transformations \cite{hafner2023mastering, webber2012bi}. Conversely, HarmonyDream introduces a dynamic approach to loss scaling within world model learning, optimizing multi-task learning efficiency through an intricate balance of scale, dimensionality, and training dynamics \cite{ma2023harmony}. The synergistic integration of DreamerV3's symlog transformation with HarmonyDream's dynamic loss adjustment holds the potential to further elevate world models' performance and versatility.

The Image-based Joint-Embedding Predictive Architecture (I-JEPA) \cite{assran2023self} illustrates an approach for learning highly semantic image representations without relying on hand-crafted data augmentations. I-JEPA predicts missing target information using abstract representations, effectively eliminating unnecessary pixel-level details. This enables the model to learn more semantic features, achieving more accurate analysis and completion of incomplete images through self-supervised learning of the world’s abstract representations. Beyond images, this architecture demonstrates high scalability with the Audio-based Joint-Embedding Predictive Architecture (A-JEPA) \cite{fei2023jepa}, setting new state-of-the-art performance on multiple audio and speech classification tasks, outperforming models that rely on externally supervised pre-training.

In robotic manipulation, such as Fetch \cite{plappert2018multi}, DeepMind Control Suite \cite{tassa2018deepmind}, and Meta-world \cite{yu2020meta}, the Latent Explorer Achiever (LEXA) \cite{mendonca2021discovering} outperforms previous unsupervised methods in 40 robot operation and movement tasks by training an explorer and  {an} achiever simultaneously through imagination. Additionally, in these tasks, L$^3$P \cite{zhang2021world} devises a novel algorithm to learn latent landmarks scattered across the goal space, achieving dominant performance in both learning speed and test-time generalization in three robotic manipulation environments. Google's team has innovatively applied the concept of world models to robot navigation tasks, utilizing them to acquire information about the surrounding environment and enable the intelligent agent to predict the consequences of its actions in specific contexts. Pathdreamer's implementation in robot navigation leverages world models for enhanced environmental awareness and predictive planning, achieving a notable improvement in navigation success rates through its innovative use of 3D point clouds for environment representation \cite{koh2021pathdreamer}. Furthermore, SafeDreamer integrates a Lagrangian-based approach into the Dreamer framework for safety reinforcement learning, demonstrating the feasibility of high-performance, low-cost safety applications \cite{huang2023safe}.

The rapid training capabilities of world models, exemplified by DayDreamer's real-world robot learning efficiency, contrast starkly with traditional methods, highlighting the transformative potential of these models in accelerating learning processes and enhancing performance \cite{wu2023daydreamer, haarnoja2018soft}.

Virtual scene and video generation emerge as pivotal applications, with SORA and Genie leading advancements in this space. SORA's capability to produce coherent, high-definition videos from diverse prompts represents a significant step towards simulating complex world dynamics. Despite its challenges in physical interaction simulation, SORA's consistent 3D spatial representation underscores its potential as a foundational world model \cite{brooks_video_2024}. Genie's interactive environment generation, though not as advanced in video quality as SORA, introduces a novel dimension of user-driven world manipulation, offering a glimpse into future applications of world models in creating immersive, controllable virtual realities \cite{bruce2024genie}.

 {In summary, this comprehensive review in this section underscores the exceptional versatility and advancing frontiers of world models, illustrating their foundational role in driving innovation across gaming, robotics, virtual environment generation, and beyond.} The convergence of these models' capabilities with dynamic adaptation and multi-domain generalization heralds a new era of AI, where world models not only serve as tools for specific tasks but also as platforms for broader exploration, learning, and discovery.

\section{World Models in Autonomous Driving}
This section delves into the transformative application of world models within the autonomous driving sector, underscoring their pivotal contributions to environmental comprehension, dynamic prediction, and the elucidation of physical principles governing motion. As an emergent frontier in the application of world models, the autonomous driving domain presents unique challenges and opportunities for leveraging these advanced computational frameworks. Despite the burgeoning interest, the integration of world models into autonomous driving predominantly revolves around scenario generation and planning and control mechanisms, areas ripe for exploration and innovation.

\begin{figure*}[ht]
\centering  
\label{Fig.4}
\includegraphics[width=0.75\textwidth]{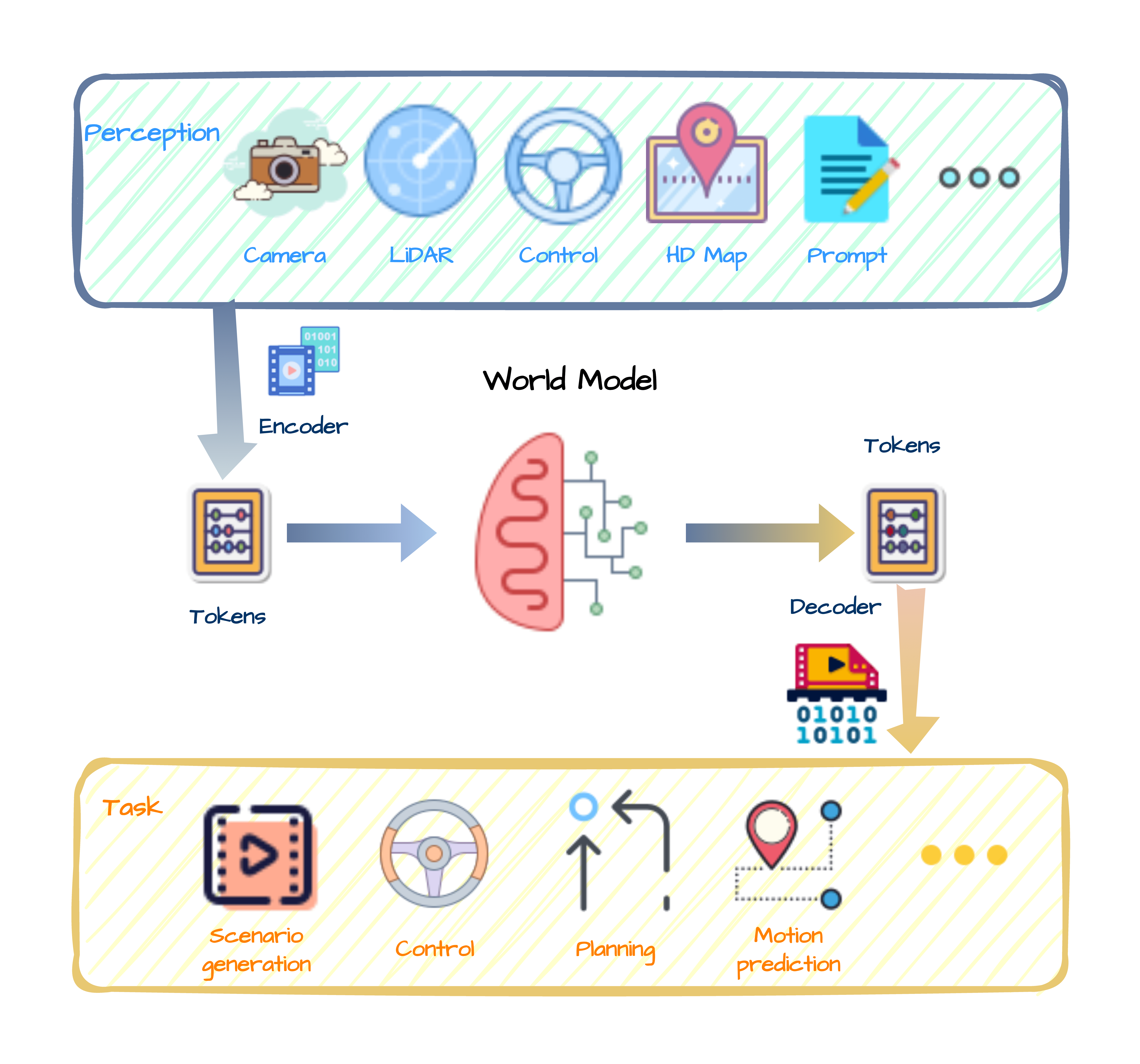}
\caption{World Models in Autonomous Driving Pipelines.}
\end{figure*}

\subsection{Driving Scenario Generation}\label{Driving scenario generation}

The acquisition of data in autonomous driving encounters substantial hurdles, including high costs associated with data collection and annotation, legal constraints, and safety considerations. World models, through self-supervised learning paradigms, offer a promising solution by enabling the extraction of valuable insights from vast quantities of unlabeled data, thereby enhancing model performance cost-effectively. The application of world models in driving scenario generation is particularly noteworthy, as it facilitates the creation of varied and realistic driving environments. This capability significantly enriches training datasets, equipping autonomous systems with the robustness to navigate rare and intricate driving scenarios \cite{tian2022parallel}.

GAIA-1 \cite{hu2023gaia} represents a novel autonomous generative AI model capable of creating realistic driving videos using video, text, and action inputs. Trained on extensive real-world driving data from British cities via Wayve, GAIA-1 learns and understands some real-world rules and key concepts in driving scenarios, including different types of vehicles, pedestrians, buildings, and infrastructure. It can predict and generate subsequent driving scenarios based on a few seconds of video input. Notably, the generated future driving scenarios are not closely tied to the prompt video but are based on GAIA-1's understanding of world rules. Employing an autoregressive transformer network at its core, GAIA-1 predicts upcoming image tokens conditioned on the input image, text, and action tokens, and then decodes these predictions back to pixel space. GAIA-1 can predict multiple potential futures and generate diverse videos or specific driving scenarios based on prompts (e.g., changing weather, scenes, traffic participants, vehicle actions), even including actions and scenes beyond its training set (e.g., forced entry onto sidewalks). This demonstrates its ability to understand and infer driving concepts not present in its training set,  {which also proves its capabilities of counterfactual reasoning.} In the real world, such driving behaviors are hard to acquire data for due to their riskiness. Driving scenario generation allows for simulated testing, enriching data composition, enhancing system capabilities in complex scenarios, and better evaluating existing driving models. Moreover, GAIA-1 generates coherent actions and effectively captures the perspective influences of 3D geometric structures, showcasing its understanding of contextual information and physical rules.  {Combined with its demonstrated ability for counterfactual reasoning, it can be said that GAIA-1 marks a high level of achievement in the world models of autonomous driving, both in the understanding of abstract concepts and in causal reasoning.}

DriveDreamer \cite{wang2023drivedreamer}, also dedicated to driving scenario generation, differs from GAIA-1 as it is trained on the nuScenes dataset \cite{caesar2020nuscenes}. Its model inputs include  {more} elements like HD maps and 3D boxes, allowing for more precise control over driving scenario generation and deeper understanding, thus improving video generation quality. Additionally, DriveDreamer can generate future driving actions and corresponding predictive scenarios, aiding in decision-making.

\begin{figure*}[ht]
\centering  
\includegraphics[width=1\textwidth]{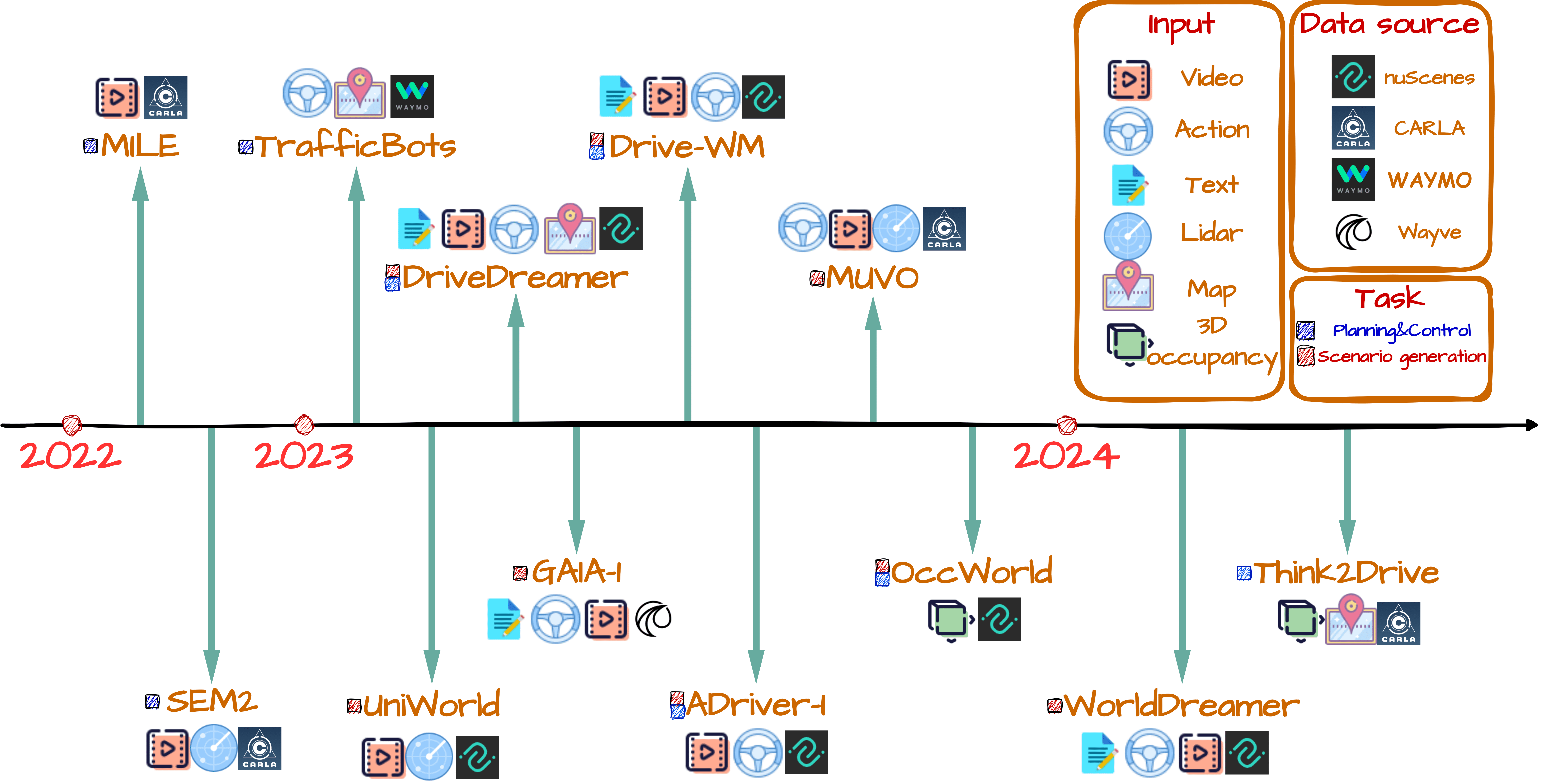}
\caption{Chronological Overview of World Models in Autonomous Driving.}
\label{Fig.5}
\end{figure*}

ADriver-I employs current video frames and historical vision-action pairs as inputs for a multimodal large language model (MLLM) \cite{chiang2023vicuna,touvron2023llama} and a video latent diffusion model (VDM) \cite{rombach2022high}. The MLLM outputs control signals in an autoregressive manner, which serve as prompts for VDM to predict subsequent video outputs. Through continuous prediction cycles, ADriver-I achieves infinite driving in the predictive world.  {In ADricer-I, the combination of world model and MLLM significantly improves the interpretability of prediction and decision-making, and also indicates the feasibility of combining the world model as a fundamental model with other models.}

Drawing inspiration from the success of large language models, WorldDreamer \cite{wang2024worlddreamer} approaches world modeling as an unsupervised visual sequence modeling challenge. It utilizes the STPT to concentrate attention on local patches within spatiotemporal windows. This focus facilitates dynamic learning of visual signals and accelerates the convergence of the training process. Although World Dreamer is a general-purpose video generation model, it has demonstrated exceptional performance in generating autonomous driving videos.

Beyond visual information, driving scenarios also include a plethora of critical physical data. MUVO \cite{bogdoll2023muvo} leverages the world model framework for the prediction and generation of driving scenes and integrates both LIDAR point clouds and visual inputs to predict videos, point clouds, and 3D occupancy grids of future driving scenes. Such a comprehensive approach significantly enhances the quality of predictions and generated outcomes. Particularly, the outcome of 3D occupancy grids can be directly applied to downstream tasks. Taking it a step further, OccWorld \cite{zheng2023occworld} and Think2Drive \cite{li2024think2drive} directly utilize the 3D occupancy information as system inputs to predict the evolution of the surrounding environment and plan the actions of the autonomous vehicles.  {It's evident that as research progresses, world model studies for scenario generation in the autonomous driving domain gradually evolve towards a multi-modal approach. World models have demonstrated a versatile capability in processing multi-modal information.}

\subsection{Planning and Control}\label{Planning and control}

Beyond scenario generation, world models are instrumental in learning within driving contexts, evaluating potential futures, and refining planning and control strategies. For instance, Model-based Imitation LEarning (MILE) \cite{hu2022model} adopts a model-based imitation learning approach to jointly learn the dynamics model and driving behavior in CARLA from offline datasets. MILE employs a 'generalized inference algorithm' for rational and visualizable imagination and prediction of future driving environments, using imagination to compensate for missing perceptual information. This ability enables planning future actions, allowing autonomous vehicles to operate without high-definition maps. In inexperienced test scenarios within the CARLA simulator, MILE significantly outperformed the state-of-the-art models, improving the Driving Score from 46 to 61 (compared to an expert data score of 88). MILE is characterized by long temporal and highly diversified future predictions. Using a decoder on the predicted future states, MILE demonstrates stable driving across various scenarios.

SEM2 \cite{gao2022enhance}, building upon RSSM, introduces the semantic masked world model to enhance the sampling efficiency and robustness of end-to-end autonomous driving. The authors contend that world models' latent states contain too much task-irrelevant information, adversely affecting sampling efficiency and system robustness. Moreover, due to imbalanced training data, world models struggle to handle unexpected situations. To address these issues, a signature filter is introduced to extract key task features, reconstructing semantic masks using the filtered features. For data imbalance, a sampler is used to balance the data distribution.  {In each batch of training, samples from various scenarios are evenly added, to achieve an even and balanced distribution of training samples, which is conducive to generalization and solving corner cases.} After being trained and tested in CARLA, the performance of SEM2 showed substantial improvement over DreamerV2.

Considering that most autonomous vehicles typically have multiple cameras, multi-view modeling is also a crucial aspect of world models. Drive-WM \cite{wang2023driving} is the first multi-view world model designed to enhance the safety of end-to-end autonomous driving planning. Drive-WM, through multi-view and temporal modelling, jointly generates frames for multiple views and then predicts intermediate views from adjacent ones, significantly improving consistency across multiple views. Additionally, Drive-WM introduces a simple unified conditional interface, flexibly applying images, actions, text, and other conditions, simplifying the conditional generation process. Trained and validated on the nuScenes dataset \cite{caesar2020nuscenes} with six views, Drive-WM selects the best trajectory by sampling predicted candidate trajectories and using an image-based reward function. Furthermore, consistent with GAIA-1, Drive-WM's ability to navigate in non-drivable areas showcases the world model's understanding and potential in handling out-of-domain cases. Besides, drawing inspiration from the seminal work of Alberto Elfes \cite{elfes1989using}, UniWorld \cite{min2023uniworld} introduces an innovative approach by utilizing multi-frame point cloud fusion as the ground truth for generating 4D occupancy labels. This method accounts for the temporal-spatial correlations present in images from multi-camera systems. By leveraging unlabeled image-lidar pairs, UniWorld undergoes pre-training for world models, significantly enhancing the understanding of environmental dynamics. When tested on the nuScenes dataset, UniWorld demonstrates notable improvements in IoU for tasks such as motion prediction and semantic scene completion compared to methods that rely on monocular pre-training.

TrafficBots \cite{zhang2023trafficbots}, also an end-to-end autonomous driving model, places a greater emphasis on predicting the actions of individual agents within a scene. Conditioning on each agent's destination, TrafficBots employs a Conditional Variational Autoencoder (CVAE) \cite{sohn2015learning} to learn distinct personalities for each agent, thereby facilitating action prediction from a BEV perspective. Compared to alternative approaches, TrafficBots offers faster operation speeds and can scale to accommodate more agents. Although its performance may not yet rival state-of-the-art open-loop strategies, TrafficBots showcases the potential of closed-loop strategies for action prediction.

\subsection{Conclusion}
 {
Fig. \ref{Fig.5} represents a chronological overview of existing world models in the field of autonomous driving, including inputs, tasks, and training datasets. Since employing world models in the autonomous driving sector remains a nascent topic, different world models' operational tasks and input-output mechanisms vary considerably. In the domain of scenario generation, this not only encompasses the generation of predicted scenario videos but also includes subdivisions such as scenario information completion and 3D occupancy prediction, among others. In the control domain, it involves autonomous driving based on input from sensors, vehicle control based on prompt words, and more. Additionally, it can be integrated with scenario generation, outputting predicted scenarios corresponding to control info, thereby offering a pathway to enhance the interpretability of autonomous driving systems.
}

 {
For those limitations, comparing the performance of different world models faces numerous challenges, including variations in tasks, validation datasets, and the criteria used for measuring performance. For world models in the context of scenario video generation using the nuScenes dataset, they employ FID (Fréchet Inception Distance) and FVD (Fréchet Video Distance) as metrics for assessing video quality. These metrics are critical for understanding how closely the generated scenes resemble real-world scenarios captured in the dataset, with lower scores indicating higher similarity and, therefore, better model performance. According to the comparison result of FID and FVD, we can find that Adriver-I has better video quality than DriveWM and DriveDreamer.
}

 {
For other world models, although a direct horizontal comparison might not be feasible due to the diversity of tasks and performance metrics, they have achieved state-of-the-art results in their tasks, compared with the traditional methods. 
}

\section{Challenges and Future Perspectives}\label{Challenges and future perspectives}

The advancement of world models in the realm of autonomous driving presents a frontier of innovation with the potential to redefine vehicular mobility. However, this promising landscape is not without its challenges. Addressing these hurdles and exploring future perspectives necessitates a deep dive into both the technical complexities and the broader societal implications.

\subsection{Technical and Computational Challenges}
\paragraph{Long-Term Scalable Memory Integration}  The quest to imbue world models with long-term, scalable memories that mirror the intricacies of human cognitive processes remains a formidable challenge in the realm of autonomous driving.
Contemporary models grapple with issues such as vanishing gradients \cite{pascanu2013difficulty} and catastrophic forgetting \cite{kirkpatrick2017overcoming}, which severely limit their long-term memory capabilities. Transformer architectures, despite their advancements in facilitating access to historical data through self-attention mechanisms, encounter obstacles in scalability and speed when processing lengthy sequences. Innovative approaches, exemplified by studies like TRANSDREAMER \cite{chen2022transdreamer} and S4WM \cite{deng2024facing}, explore alternative neural architectures aiming to surmount these barriers. Notably, S4WM has demonstrated superior performance in maintaining high-quality generation over sequences up to 500 steps, markedly surpassing traditional architectures. Yet, the decline in performance observed beyond 1000 steps accentuates the existing disparity between the capacities of artificial and biological memory systems.

To bridge this gap, future research endeavors may pivot towards a multi-pronged strategy encompassing the augmentation of network capacities, the integration of sophisticated external memory modules, and the exploration of iterative learning strategies. These efforts aim not only to extend the temporal reach of memory in world models but also to enhance their ability to navigate the complex decision-making processes inherent in autonomous driving. By fostering a deeper synergy between computational efficiency and memory scalability, these advancements could significantly propel the capabilities of autonomous vehicles, enabling them to adapt and respond to the ever-changing dynamics of real-world driving environments with unprecedented precision and reliability.

\paragraph{Simulation-to-Real-World Generalization} The disparity between simulated training environments and the multifaceted nature of real-world conditions presents a critical bottleneck in the evolution of autonomous driving technologies. Current simulation platforms, while advanced, fall short of perfectly mirroring the unpredictability and variability of real-world scenarios. This discordance, manifesting in discrepancies in physical properties, sensor noise, and the occurrence of unforeseen events, critically undermines the applicability of world models trained solely in simulated environments \cite{stocco2022mind}. 

 {Although this challenge is still inevitable due to technological limitations, the endeavour to develop world models capable of seamless generalization from simulation to real-world driving scenarios is paramount. This requires not only the refinement of simulation technologies to capture the subtleties and unpredictability of real-world environments more accurately but also the development of models that are inherently robust to the variances between simulated and real-world data. 
Moreover, the integration of advanced sensory fusion techniques and the exploration of novel learning paradigms, such as meta-learning and reinforcement learning from diverse data sources, could further empower world models to adapt dynamically to the complexities of real-world driving. These advancements are pivotal for the realization of truly autonomous driving systems capable of navigating the myriad challenges posed by real-world environments with agility, accuracy, and safety.}

\paragraph{Theory and Hardware Breakthroughs}  {World models currently excel in generative tasks more than in pure predictive tasks, such as motion prediction. This is partly because these models still fall short of perfectly mimicking the evolution of the real world, including the balance between determinism and randomness. Additionally, after processing through sensors and encoders, the information entering the latent space loses a significant amount of detail compared to the real world. Although it's possible to ignore some details and focus on the main factors likely to impact the future through attention mechanisms or special model structures—thus reducing data volume and avoiding irrelevant detail interference—neglecting these nuances inevitably leads to performance losses, creating a bottleneck for the model's predictive capacity. Addressing and restoring sufficient detail, on the other hand, poses a significant challenge to memory and computational power. Despite the encouraging achievements of world models, these newly developed models still require further refinement in theory and structure to serve as foundational AI models. Consequently, transitioning world models from simulators to real-world deployment is expected to take a considerable amount of time.}

 {In summary, while continuously improving the theory of world models, advancements in hardware are also necessary. This includes lossless collection and processing of multimodal information, along with enhancements in the computational capabilities of hardware facilities.}

\subsection{Ethical and Safety Challenges}
\paragraph{Decision-Making Accountability} The imperative of ensuring accountability within the autonomous decision-making frameworks of vehicles stands as a paramount ethical concern, necessitating the development of systems characterized by an unparalleled level of transparency. The complexity inherent in the algorithms guiding autonomous vehicles necessitates a mechanism that not only facilitates decision-making in critical and routine scenarios but also enables these systems to articulate the rationale underpinning their decisions. This transparency is vital for building and maintaining trust among end-users, regulatory bodies, and the broader public.

To achieve this, there is a pressing need for the integration of explainable AI (XAI) principles directly into the development of world models \cite{gade2019explainable}. XAI aims to make AI decisions more interpretable to humans, providing clear, understandable explanations for the actions taken by autonomous vehicles. This involves not merely an exposition of the decision-making process but a comprehensive delineation of the ethical, logical, and practical considerations influencing these decisions. Implementing XAI within autonomous driving systems necessitates a multidisciplinary approach, drawing on expertise from AI development, ethics, legal standards, and user experience design \cite{ignatiev2020towards}.  {In this respect, the regulatory guidance that has emerged regarding explainable automated decision-making under the General Data Protection Regulation (GDPR) should be of assistance.}

\paragraph{Privacy and Data Integrity}
The reliance on autonomous driving technologies on extensive datasets for operation and continuous improvement brings to the forefront significant concerns regarding privacy and data security.  {For companies developing autonomous driving systems, the vehicle-related data collected, including information about passengers, origins and destinations, travel times, and routes, constitutes both a legally obtained and highly valuable commercial asset \cite{collingwood2017privacy}.} 
The safeguarding of personal information against unauthorized access and breaches is a critical priority, requiring a robust framework for the ethical handling and protection of data.

Addressing these concerns involves a multifaceted strategy that extends beyond compliance with existing privacy regulations such as GDPR in Europe  {or California Consumer Privacy Act (CCPA) in the US \cite{regulation2018general}}. It entails the establishment of stringent data governance policies that dictate the collection, processing, storage, and sharing of data. These policies should be designed to minimize data exposure and ensure data minimization principles, whereby only the data necessary for specific legitimate purposes is processed. Moreover, the deployment of advanced cybersecurity measures is critical to protect data integrity and confidentiality. This includes the utilization of encryption technologies, secure data storage solutions, and regular security audits to identify and mitigate potential vulnerabilities. Additionally, fostering transparency with users about how their data is collected, used, and protected is fundamental \cite{maurer2016autonomous}. This can be achieved through clear, accessible privacy policies and mechanisms that empower users with control over their personal information, including options for data access, correction, and deletion \cite{bonnefon2016social}. 

\paragraph{Responsibility and Criterion}

 {As world models support or take over driving tasks in autonomous driving systems, human responsibility is not lessened or eliminated but rather redistributed among the network of individuals and organizations involved in their creation, deployment, and usage. This shift demands different requirements from the participants, calling for new research and policies to govern this transformation in demand.}

 {Obligation-wise, policymakers should encourage research activities in ethics design, aiming to establish it as a solid academic field, akin to medical ethics \cite{veliz2019three}. Moreover, there is a need to create suitable educational environments to promote citizen education on the obligations of various stakeholders. Ethically, policymakers, manufacturers, and deployers should set up mechanisms to reward those who actively take on responsibilities within organizations or professional associations focused on ethical design and deployment, thus driving a shift in culture \cite{schiebinger2008gendered}.}

 {Given the complexity of the socio-technical systems autonomous driving systems belong to, traditional, moral, and legal standards for attributing responsibility to individual human agents may not easily apply to behaviors emerging from interactions between humans and intelligent systems. Policymakers, in collaboration with researchers, manufacturers, and deployers, should establish clear and fair legal rules for assigning liability in case of incidents. This could involve developing new insurance models. Such rules must find a balance between avoiding culpability gaps and fulfilling the demands of corrective justice \cite{bonnefon2020ethics}.}

\subsection{Future Perspectives}
\paragraph{Bridging Human Intuition and AI Precision}
One groundbreaking perspective is the evolution of world models towards facilitating a cognitive co-piloting framework within autonomous vehicles. Unlike traditional autonomous systems, which rely solely on pre-defined algorithms and sensor inputs for decision-making, cognitive co-piloting aims to blend the nuanced, intuitive decision-making capabilities of human drivers with the precision and reliability of AI. By leveraging advanced world models, vehicles can gain an unprecedented level of environmental awareness and predictive capability, mirroring human cognitive processes such as anticipation, intuition, and the ability to navigate complex socio-technical environments.

This integration enables autonomous vehicles to not only react to the immediate physical world but also to understand and adapt to the social and psychological dimensions of driving—interpreting gestures, predicting human behaviors, and making decisions that reflect a deeper understanding of human norms and expectations. For instance, the Context-Aware Visual Grounding (CVAG) model integrated with GPT-4 can learn human emotional characteristics through contextual semantics and effectively process and interpret cross-modal inputs to assist recognition and execution decisions \cite{liao2024gpt}. Likewise, a world model equipped with cognitive co-piloting capabilities could accurately predict pedestrian movements in urban settings, navigate social driving conventions at four-way stops, or adapt driving styles in response to passenger comfort and feedback.

\paragraph{Harmonizing Vehicles with the Urban Ecosystem} Another visionary perspective involves the role of world models in transforming autonomous vehicles into agents of ecological engineering, harmonizing with urban ecosystems through adaptive, responsive behaviors that contribute to environmental sustainability. World models, with their deep understanding of complex systems and dynamics, can enable autonomous vehicles to optimize routes and driving patterns not just for efficiency and safety but also for environmental impact, such as minimizing emissions, reducing congestion, and promoting energy conservation \cite{zhu2022flow,xu2024integrating,shen2023coastal}.

 {A global survey \cite{goddard2021global} reveals that over 60\% of respondents believe that with the advancement in urban system automation, the pollution generated by transportation systems and the likelihood of vehicle collisions will decrease. Furthermore, more than 70\% of the respondents anticipate improvements in traffic noise. Meanwhile, those who believe that roadworks and transport system management will see enhancements exceed 80\%. This data reflects a widespread optimism towards the potential environmental and safety benefits of increasingly automated urban transport systems. World models represent a crucial direction for development that can facilitate a higher degree of automation in vehicles and traffic systems, showcasing their significance in advancing urban infrastructure towards safer and more sustainable futures.}

\section{Conclusion}
In conclusion, this survey has delved into the transformative potential of world models in the autonomous driving landscape, highlighting their pivotal role in advancing vehicle autonomy through enhanced prediction, simulation, and decision-making capabilities. Despite the significant progress documented, challenges such as long-term memory integration, simulation-to-real-world generalization, and ethical considerations underscore the complexity of deploying these models in real-world applications. Addressing these challenges necessitates a multidisciplinary approach that combines advancements in AI research with ethical frameworks and innovative computational solutions. Looking ahead, the evolution of world models promises to not only enhance autonomous driving technologies but also to redefine our interaction with automated systems, underscoring the need for continued research and collaboration across fields. As we stand on the cusp of this technological frontier, it is imperative that we navigate the ethical implications and societal impacts with diligence and foresight, ensuring that the development of autonomous driving technologies remains aligned with broader societal values and safety standards.


%

\section*{Acknowledgment}
This research is supported by the Science and Technology Development Fund of Macau SAR (Project no.: 0021/2022/ITP, 001/2024/SKL).

\ifCLASSOPTIONcaptionsoff
  \newpage
\fi

\printbibliography
%

\begin{IEEEbiography}
[{\includegraphics[width=1in,height=1.40in, clip,keepaspectratio]{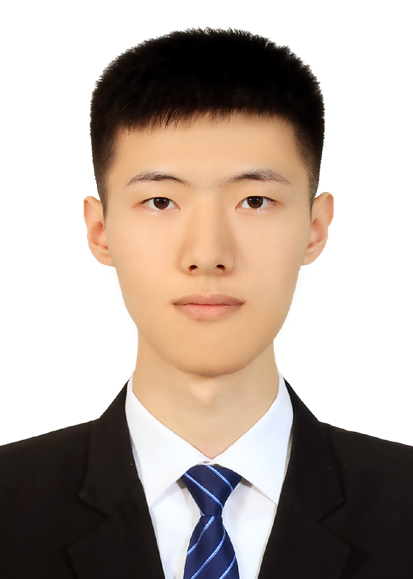}}]{Yanchen Guan} is currently a PhD student at the State Key Laboratory of Internet of Things for Smart City and the Department of Civil Engineering at the University of Macau. He holds a master’s degree in Mobility Engineering from Politecnico di Milano (2023) and a bachelor’s degree in Mechatronics Engineering from Harbin Institute of Technology (2019). His research primarily focuses on autonomous driving, intelligent transportation systems, mechanical structures, and data analysis.
\end{IEEEbiography}

\begin{IEEEbiography}
[{\includegraphics[width=1.1in,height=1.40in, clip,keepaspectratio]{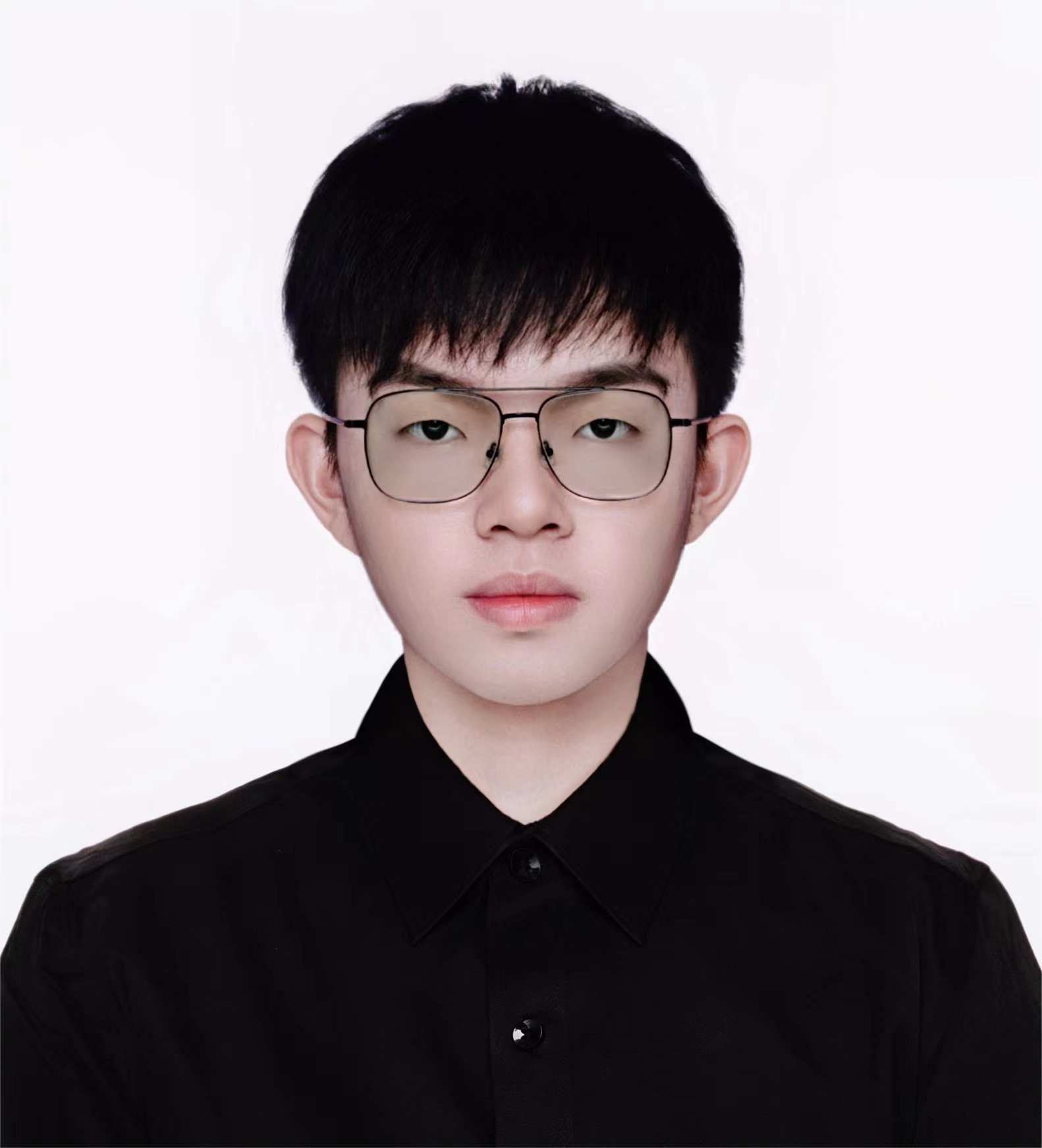}}]{Haicheng Liao} (Student Member, IEEE) received the B.S. degree in software engineering from the University of Electronic Science and Technology of China (UESTC) in 2022. He is currently pursuing the Ph.D. degree at the State Key Laboratory of Internet of Things for Smart City and the Department of Computer and Information Science, University of Macau. His research interests include connected autonomous vehicles and the application of deep reinforcement learning to autonomous driving.
\end{IEEEbiography}


\begin{IEEEbiography}
[{\includegraphics[width=1in,height=1.25in,clip,keepaspectratio]{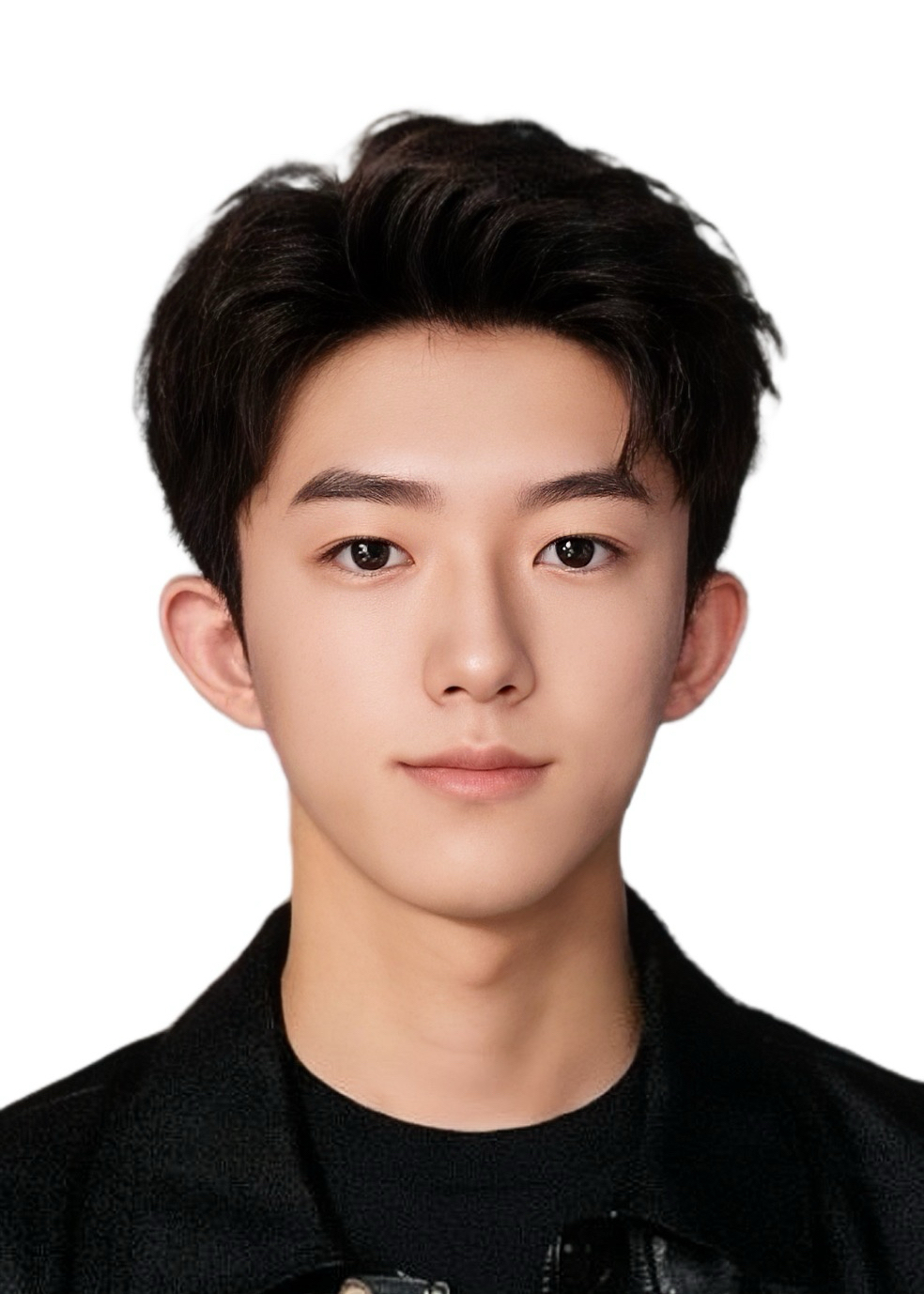}}] {Zhenning Li} (Member, IEEE) received his Ph.D. in Civil Engineering from the University of Hawaii at Manoa in 2019. Currently, he holds the position of Assistant Professor at the State Key Laboratory of Internet of Things for Smart City, as well as the Departments of Civil and Environmental Engineering and Computer and Information Science at the University of Macau, Macau SAR.  He has published over 50 papers in top journals and conferences. His main areas of research focus on the intersection of connected autonomous vehicles and Big Data applications in urban transportation systems. He has been honored with several awards, including the TRB Best Young Researcher award and the CICTP Best Paper Award, amongst others.
\end{IEEEbiography}

\begin{IEEEbiography}
[{\includegraphics[width=1in,height=1.25in,clip,keepaspectratio]{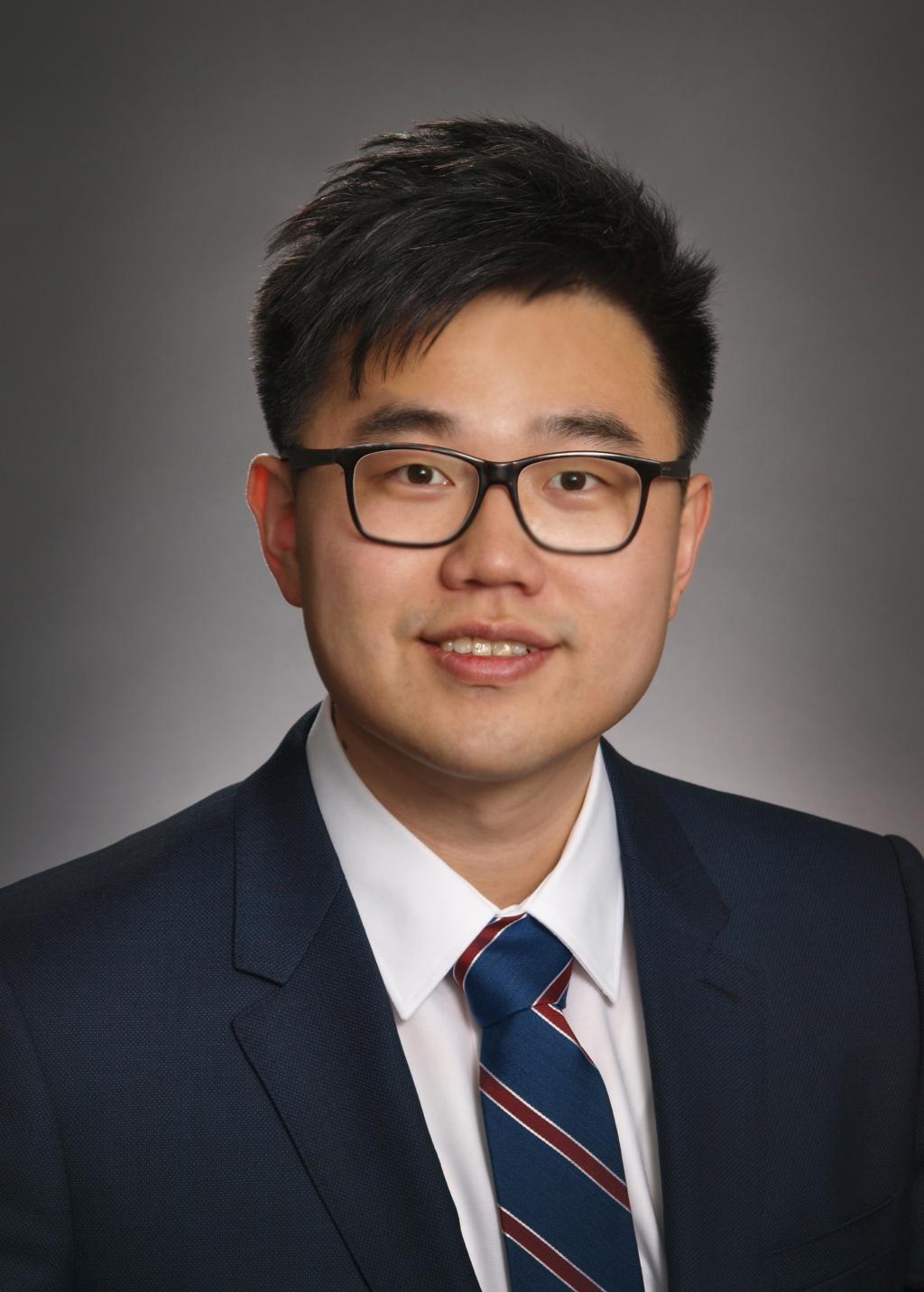}}] {Jia Hu} (Member, IEEE) is currently working as a Zhongte  Distinguished Chair of Cooperative Automation with the  College of Transportation Engineering, Tongji University.  Before joining Tongji University, he was a Research  Associate with the Federal Highway Administration  (FHWA), USA. He is an Editorial Board Member of the  Journal of Intelligent Transportation Systems and the  International Journal of Transportation Science and  Technology. He is a member of TRB (a Division of the  National Academies) Vehicle Highway Automation  Commit-tee, the Freeway Operations Committee, the Simulation subcommittee of the Traffic Signal Systems Committee, and the Advanced Technologies Committee of the ASCE Transportation and Development Institute. He is the Chair of the  Vehicle Automation and Connectivity Committee of the World Transport  Convention. He is an Associate Editor of the American Society of Civil  Engineers Journal of Transportation Engineering and IEEE OPEN JOURNAL  OF INTELLIGENT TRANSPORTATION SYSTEMS.
\end{IEEEbiography}

\begin{IEEEbiography}
[{\includegraphics[width=1in,height=1.25in,clip,keepaspectratio]{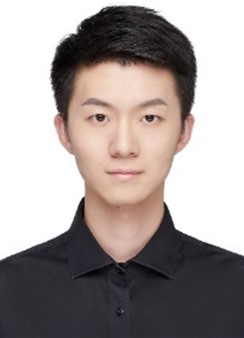}}]
{Runze Yuan} received a bachelor's degree from Tsinghua University, Beijing, in 2017. He received the PhD degree from the University of Hawaii at Manoa, Honolulu, in 2023. He is currently a postdoc in the Department of Automation at Tsinghua University, Beijing. His main research areas include microscopic traffic flow modeling, artificial intelligence in transportation, traffic flow detection and control, and intelligent transportation systems.
\end{IEEEbiography}

\begin{IEEEbiography}
[{\includegraphics[width=1in,height=1.25in,clip,keepaspectratio]{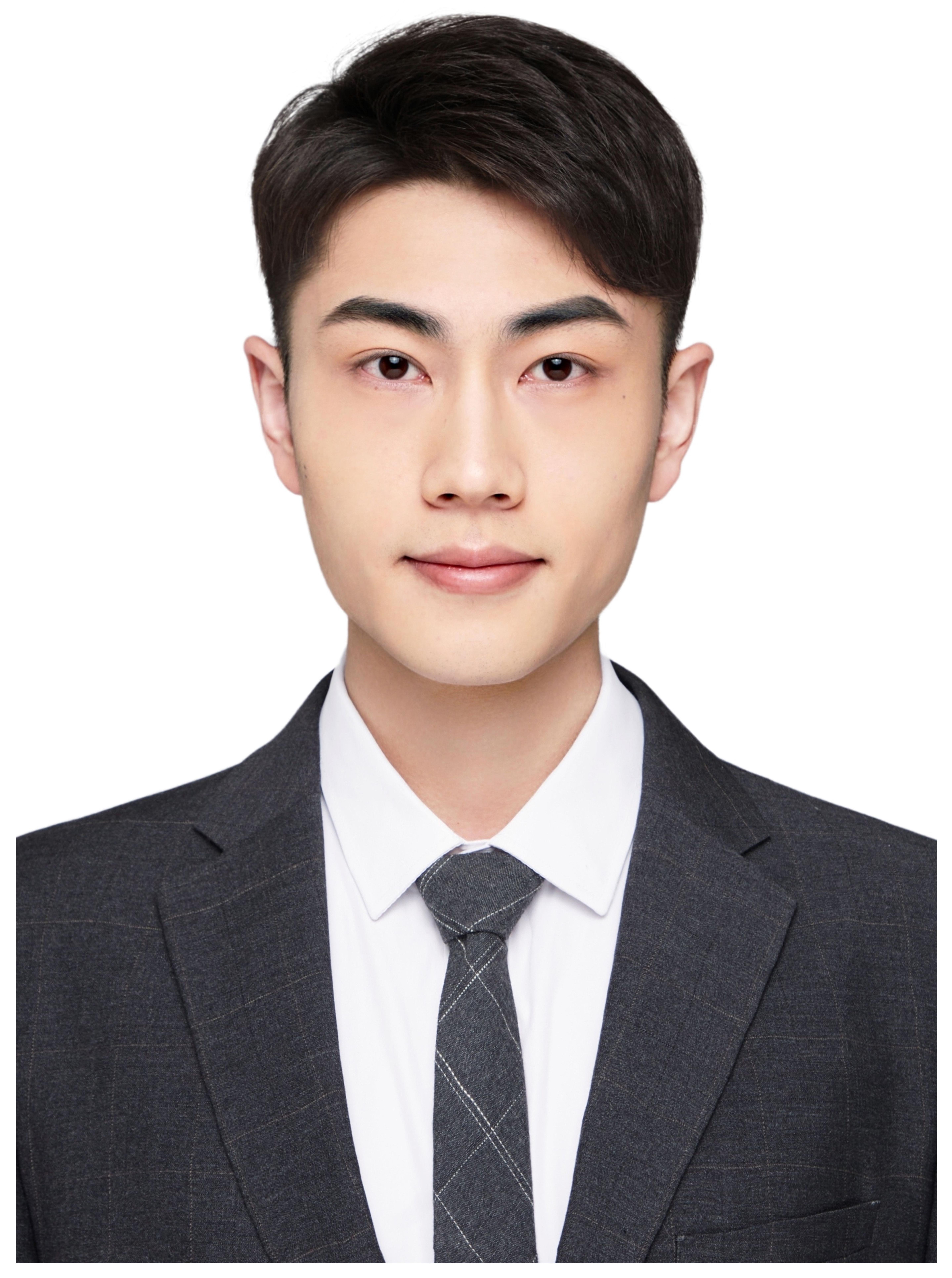}}]
{Yunjian Li} earned his Ph.D. in Applied Physics and Materials Engineering from the University of Macau in 2023. Presently, he serves as an Assistant Professor in the Faculty of Innovation Engineering at Macau University of Science and Technology. He has authored 15 SCI papers published in prestigious journals such as Nature Communications and Cement and Concrete Research.
\end{IEEEbiography}

\begin{IEEEbiography}
[{\includegraphics[width=1in,height=1.25in,clip,keepaspectratio]{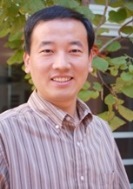}}]
{Guohui Zhang} is currently a Professor with the  Department of Civil and Environmental Engineering, University of Hawai'i at Manoa. His primary research interests include transportation system resilience, transportation systems modeling, planning, and operation, traffic sensing and sensor data analytics, artificial intelligence in transportation,  connected and autonomous vehicle systems, and transportation safety and security. He serves as a member of the American Society of Civil Engineers Connected and Autonomous Vehicle Impact  Committee and the Transportation Research Board Information Systems and  Technology Committee and as a panelist for multiple National Collaborative  Highway Research Program projects.
\end{IEEEbiography}

\begin{IEEEbiography}
[{\includegraphics[width=1in,height=1.25in,clip,keepaspectratio]{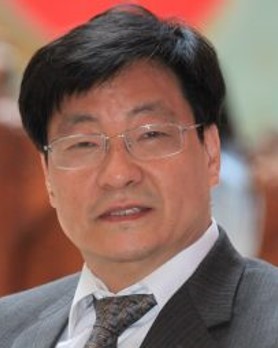}}]{Chengzhong Xu} (Fellow, IEEE) received the Ph.D. degree from The University of Hong Kong, in 1993. He is currently the chair professor of computer science and the dean with the Faculty of Science and Technology, University of Macau. Prior to this, he was with the faculty at Wayne State University, USA, and the Shenzhen Institutes of Advanced Technology, Chinese Academy of Sciences, China. He has published more than 400 papers and more than 100 patents. His research interests include cloud computing and data-driven intelligent applications. He was the Best Paper awardee or the Nominee of ICPP2005, HPCA2013, HPDC2013, Cluster2015, GPC2018, UIC2018, and AIMS2019. He also won the Best Paper award of SoCC2021. He was the Chair of the IEEE Technical Committee on Distributed Processing from 2015 to 2019.
\end{IEEEbiography}



\end{document}